\documentclass{article}
\usepackage{lmodern}
\usepackage[preprint]{colm2026_conference}

\usepackage{microtype}
\usepackage{hyperref}
\usepackage{url}
\usepackage{booktabs}

\usepackage{amsmath}
\usepackage{amsfonts}
\usepackage{amsthm}
\usepackage{siunitx} 
\usepackage[dvipsnames]{xcolor}
\usepackage{colortbl}
\usepackage[most]{tcolorbox}
\usepackage{multirow, tabularx}
\usepackage{wrapfig}
\usepackage{pifont}
\usepackage{fontawesome5} 
\usepackage{titletoc} 
\usepackage{caption}
\usepackage{lineno}
\newcommand{\ourBenchmark}{\textsc{Stargazer}}
\definecolor{stat-bg}{HTML}{DBEAFE}
\definecolor{phys-bg}{HTML}{FEE2E2}
\definecolor{easy-bg}{HTML}{EFF6FF}
\definecolor{med-bg}{HTML}{DBEAFE}
\definecolor{hard-bg}{HTML}{BFDBFE}
\definecolor{real-bg}{HTML}{FEE2E2}
\definecolor{skill-easy-bg}{HTML}{E0F7F6}
\definecolor{skill-med-bg}{HTML}{B2EBEA}
\definecolor{skill-hard-bg}{HTML}{81D8D0}
\newcommand{\best}[1]{\textbf{#1}}
\newcommand{\second}[1]{\underline{#1}}

\definecolor{pantone655}{HTML}{002554}
\definecolor{pantone655bg}{HTML}{EAF0F7}
\definecolor{skillboxbg}{HTML}{F4F9FF}
\definecolor{skillboxframe}{HTML}{7FB3FF}
\definecolor{skillboxtitle}{HTML}{2C6ECB}
\definecolor{cs-green}{HTML}{2E7D32}
\definecolor{cs-red}{HTML}{C62828}
\definecolor{soft-gray}{HTML}{F7F8FA}
\definecolor{soft-line}{HTML}{D7DCE2}
\definecolor{insight-bg}{HTML}{FFF8E1}
\definecolor{insight-frame}{HTML}{F9A825}
\definecolor{ok-green}{HTML}{43A047}
\definecolor{fail-red}{HTML}{E53935}

\newcommand{\cmark}{\ding{51}}
\newcommand{\xmark}{\ding{55}}
\newcommand{\diagpass}[1]{\textcolor{ok-green}{\cmark\,#1}}
\newcommand{\diagfail}[1]{\textcolor{fail-red}{\xmark\,#1}}

\newtcolorbox{insightbox}{%
  enhanced, breakable,
  colback=insight-bg, colframe=insight-frame,
  boxrule=0.8pt, arc=3pt,
  left=6pt, right=6pt, top=4pt, bottom=4pt,
  fontupper=\small,
  before upper={\textbf{Takeaway.}\;}
}
\newtcolorbox{stepbox}[2][]{%
  enhanced, breakable,
  colback=pantone655bg, colframe=pantone655,
  boxrule=0.6pt, arc=3pt,
  left=6pt, right=6pt, top=3pt, bottom=3pt,
  fontupper=\small,
  attach boxed title to top left={yshift=-2mm, xshift=4mm},
  boxed title style={colback=pantone655, colframe=pantone655, arc=2pt,
    boxrule=0pt, left=4pt, right=4pt, top=2pt, bottom=2pt},
  title={\textbf{\textcolor{white}{#2}}}, #1
}
\newtcolorbox{failbox}[2][]{%
  enhanced, breakable,
  colback=white, colframe=cs-red,
  boxrule=0.6pt, arc=3pt,
  left=6pt, right=6pt, top=3pt, bottom=3pt,
  fontupper=\small,
  attach boxed title to top left={yshift=-2mm, xshift=4mm},
  boxed title style={colback=cs-red, colframe=cs-red, arc=2pt,
    boxrule=0pt, left=4pt, right=4pt, top=2pt, bottom=2pt},
  title={\textbf{\textcolor{white}{#2}}}, #1
}
\newtcolorbox{successbox}[2][]{%
  enhanced, breakable,
  colback=white, colframe=cs-green,
  boxrule=0.6pt, arc=3pt,
  left=6pt, right=6pt, top=3pt, bottom=3pt,
  fontupper=\small,
  attach boxed title to top left={yshift=-2mm, xshift=4mm},
  boxed title style={colback=cs-green, colframe=cs-green, arc=2pt,
    boxrule=0pt, left=4pt, right=4pt, top=2pt, bottom=2pt},
  title={\textbf{\textcolor{white}{#2}}}, #1
}

\definecolor{darkblue}{rgb}{0, 0, 0.5}
\hypersetup{colorlinks=true, citecolor=darkblue, linkcolor=darkblue, urlcolor=darkblue}

\title{\ourBenchmark{}: A Scalable Model-Fitting Benchmark Environment for AI Agents under Astrophysical Constraints}

\author{%
\textbf{Xinge Liu}\textsuperscript{1*},
\textbf{Terry Jingchen Zhang}\textsuperscript{2*}\\
\textbf{Bernhard Sch\"olkopf}\textsuperscript{3,4},
\textbf{Zhijing Jin}\textsuperscript{1,2,3},
\textbf{Kristen Menou}\textsuperscript{1}\\[1ex]
\textsuperscript{1}University of Toronto \quad
\textsuperscript{2}Vector Institute\\
\textsuperscript{3}Max Planck Institute for Intelligent Systems, T\"ubingen, Germany\\
\textsuperscript{4}ELLIS Institute T\"ubingen\\[1.5ex]
\texttt{Email: xinge.liu@mail.utoronto.ca, kristen.menou@utoronto.ca}
}

\begin{document}

\ifcolmsubmission
\linenumbers
\fi

\maketitle

\begin{abstract}
The rise of autonomous AI agents suggests that dynamic benchmark environments with built-in feedback on scientifically grounded tasks are needed to evaluate the capabilities of these agents in research work. We introduce \textsc{Stargazer}, a scalable environment for evaluating AI agents on dynamic, iterative physics-grounded model-fitting tasks using inference on radial-velocity (RV) time series data. \textsc{Stargazer} comprises 120 tasks across three difficulty tiers, including 20 real archival cases, covering diverse scenarios ranging from high-SNR single-planet systems to complex multi-planetary configurations requiring involved low-SNR analysis. Our evaluation of eight frontier agents reveals a gap between numerical optimization and adherence to physical constraints: although agents often achieve a good statistical fit, they frequently fail to recover correct physical system parameters, a limitation that persists even when agents are equipped with vanilla skills. Furthermore, increasing test-time compute yields only marginal gains, with excessive token usage often reflecting recursive failure loops rather than meaningful exploration. \textsc{Stargazer} presents an opportunity to train, evaluate, scaffold, and scale strategies on a model-fitting problem of practical research relevance today. Our methodology to design a simulation-driven environment for AI agents presumably generalizes to many other model-fitting problems across scientific domains. 
\begin{center}
\small
\faGithub{} Code: \url{https://github.com/AIPS-UofT/Stargazer}

\faGlobe{} Website: \url{https://aips-uoft.github.io/Stargazer}
\end{center}
\end{abstract}

\section{Introduction}

\begin{tcolorbox}[
  enhanced,
  frame hidden,
  interior style={left color=white, right color=white, middle color=gray!15},
  arc=0pt,
  left=12pt,right=12pt,top=6pt,bottom=6pt
]
\small\itshape
If planets are colossally abundant, as it seems they are, then perhaps life is too.\\
{\raggedleft---Michel Mayor, Nobel Laureate (2019) for discovering the first exoplanet\par}
\end{tcolorbox}

Mastering the laws of physics has long been considered a defining challenge for artificial intelligence, as the discipline of physics demands tight integration of experimental observation with theoretical derivation~\citep{wang2023nature,krenn2022understanding}.
While frontier models have made rapid progress on question answering (QA) benchmarks such as HLE~\citep{HLE} and GPQA~\citep{rein2024gpqa}, AI for scientific discovery increasingly calls for agentic, multi-step workflows that involve tool-calls, simulations, and iteratively learn from feedback over repeated attempts~\citep{shen2026sciagentgymbenchmarkingmultistepscientific,chen2024scienceagentbench,lupidi2026airsbenchsuitetasksfrontier}. However, existing agentic benchmarks largely focus on simplified environments such as equation discovery~\cite{newtonbench,koblischke2025gravitybench} that do not fully capture the complexity of real-world research. We aim to go one step further in emulating high-fidelity scientific workflow of frontline researchers in the field of astrophysics.

Exoplanet discovery is tied to one of the most critical existential questions for humanities: are we alone in the universe?
A concrete step toward answering it is to find Earth-like planets with potentially habitable environments~\citep{perryman2018exoplanet}.
Since the first exoplanet discovery in 1995~\citep{mayor1995jupiter}, radial-velocity (RV) spectroscopy has provided a robust dynamical way to indirectly detect planets even when they do not pass in front of their host star. With more than six thousand confirmed exoplanets~\citep{winn2015occurrence}, RV methods remains a cornerstone for characterizing planetary systems.

We introduce \ourBenchmark{}, a high-fidelity testbed that evaluates frontier agents on an
autonomous scientific workflow of exoplanet discovery using radial velocity (RV) methods.
RV analysis is an ideal platform for evaluating scientific agents for three reasons.
First, it demands a structured, multi-step workflow, including periodogram analysis, iterative Keplerian fitting, model selection, and submission, that cannot be short-circuited by retrieval or pattern matching.
Second, success is objectively verifiable: a proposed planetary configuration either
matches the ground truth or it does not.
Third, task complexity scales with physics grounding, allowing fine-grained
difficulty control without artificial contrivance.

\ourBenchmark{} contains an infinitely scalable data synthesis pipeline with 120 sample tasks spanning three difficulty levels, including 20 tasks
drawn from real archival stellar spectra.
We evaluated 8 frontier models and report three findings: (1) statistical fit quality does not imply physical recovery; (2) token usage does not predict performance; and (3) successful agents escalate model complexity while failed agents repeat. Bootstrapped skills help on Easy-tier tasks, but do not reliably transfer to Hard-tier tasks. Critically, these challenges highlight that our simulation-driven feedback framework addresses a fundamental bottleneck in automated inference, one that presumably generalizes to diverse model-fitting problems across scientific domains. Overall, \ourBenchmark{} represents a step towards accelerating exoplanet discovery with AI agents and offers insights for future practitioners.

\begin{figure*}[tbp]
    \centering
    \includegraphics[width=1.0\textwidth]{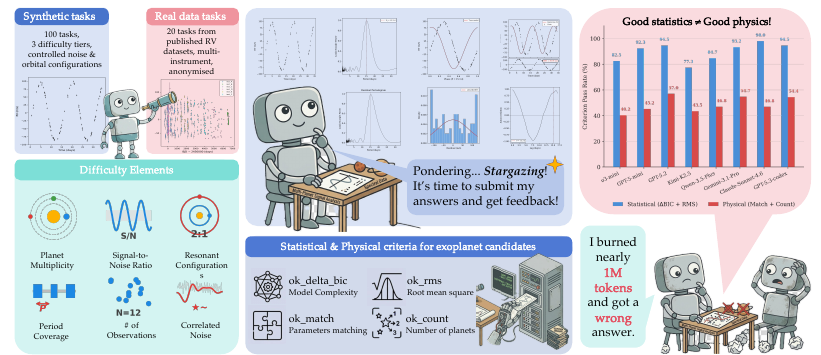}
    \vspace{-6pt}
    \caption{Overview of Stargazer. \textbf{Left:} 120 RV tasks (100 synthetic, 20 real), with synthetic difficulty controlled by six physical factors. \textbf{Center:} Agents run a periodogram-to-Keplerian workflow and are graded on statistical and physical criteria. \textbf{Right:} Models often achieve strong statistical fits but fail to recover correct orbital parameters.}
    \label{fig:overview}
\end{figure*}

\section{Related Work}

\textbf{LLM Benchmarks in Physics and Astronomy.}
Physics is gaining increasing prevalence in LLM capability evaluation.
General scientific-reasoning suites such as
\textsc{SciBench}~\citep{wang2024scibench},
\textsc{OlympiadBench}~\citep{he2024olympiadbench}, and
\textsc{GPQA}~\citep{rein2024gpqa} feature physics as a core domain,
and dedicated physics reasoning benchmarks~\citep{xu2025ugphysics, zhang2025physreason,
xiang2025seephys, zhu2025critpt, zhao2025prism} have further shown that model
performance degrades sharply as problems demand more sophisticated physical theorems and derivations.
This trend extends to astronomy, where
\textsc{AstroMLab~1}~\citep{ting2024astromlab1winsastronomy} evaluates
expert-level knowledge retrieval and
\textsc{AstroMMBench}~\citep{shi2025astrommbenchbenchmarkevaluatingmultimodal}
broadens the scope to multimodal image interpretation.
Despite this progress, existing benchmarks uniformly draw from coursework,
competitions, or factual recall, leaving open whether models can reason
over raw empirical data as a working researcher would.

\textbf{LLM Agents for Physics Research.}
Recent work explores LLM-based agents for physics research, spanning both
symbolic/theoretical reasoning and data-driven discovery.
\citet{brenner2026open} show the neuro-symbolic progress on an open theoretical
physics problem.
On the empirical side, existing benchmarks in the physical sciences focus on
narrow slices of the workflow: \textsc{AstroVisBench}~\citep{joseph2025astrovisbench}
addresses a single astronomy stage; \textsc{Gravity-Bench}~\citep{koblischke2025gravitybench}
studies gravitational-law discovery from simulation but omits scalable real
observations; and \textsc{AstroReason-Bench}~\citep{wang2026astroreasonbenchevaluatingunifiedagentic}
emphasizes mission-level planning. Complementary efforts emphasize tool use and
simulation execution under practical constraints, including \textsc{SimulCost}~\citep{cao2026simulcostcostawarebenchmarktoolkit},
\textsc{PhysicsMind}~\citep{mak2026physicsmindsimrealmechanics}, and
\textsc{SciAgentGym}~\citep{shen2026sciagentgymbenchmarkingmultistepscientific}.
More general agent benchmarks assess components that physics agents also
require, for example code generation~\citep{chen2024scienceagentbench}, data-driven scientific discovery~\citep{chen2025scienceagentbenchrigorousassessmentlanguage,lupidi2026airsbenchsuitetasksfrontier},
hypothesis search~\citep{majumder2024discoverybench}, reproducibility~\citep{siegel2024corebench},
and workflow execution~\citep{tian2024scicode}, but do not test end-to-end reasoning
from raw measurements to physical interpretation.
\ourBenchmark{} fills this gap by challenging agents to execute
end-to-end exoplanet-discovery workflows using radial-velocity methods,
requiring them to process noisy time-series data, select appropriate
physical models, and extract orbital parameters at the complexity
real-world astrophysical research demands.

\section{\ourBenchmark}

\textbf{Physics-Grounded Environment.}
\ourBenchmark{} simulates the problem of exoplanet discovery and characterization from stellar radial velocity (RV) observations. We synthesize RV time series by modeling the gravitational influence of orbiting planets on their host star. Each planetary system is parameterized by orbital period, eccentricity, argument of periastron, and orbital phase, which together determine the velocity signal induced on the star.

The stellar reflex motion is modeled using Keplerian orbital dynamics and can optionally incorporate full $N$-body integrations for multi-planet systems when dynamical interactions become significant. The resulting RV signal is sampled at irregular observation times to reflect realistic telescope schedules and observational constraints. Measurement noise and stellar activity are incorporated through Gaussian observational uncertainty and correlated noise processes.
Formally, the observed radial velocity signal at time $t$ is modeled as

\begin{equation}
v(t) = \sum_{i=1}^{N_p} v_i(t;\theta_i) + \gamma + \epsilon(t),
\end{equation}

where $N_p$ denotes the number of planets in the system, $v_i(t;\theta_i)$ represents the Keplerian velocity contribution of planet $i$ with orbital parameters $\theta_i$, $\gamma$ is the systemic velocity offset of the star, and $\epsilon(t)$ represents measurement and stellar noise. For multi-instrument datasets, a separate offset     
  $\gamma_j$ is fitted per instrument. The agent is given access only to the observed time series and must infer the underlying planetary configuration.

\subsection{Task Construction}
\label{sec:task_construction}

\ourBenchmark{} comprises 100 synthetic and 20 real-data RV tasks, grouped into three difficulty tiers: Easy (20 tasks), Medium (40 tasks), and Hard (40 tasks).
As illustrated in Figure~\ref{fig:environment_overview} (left), each synthetic task is fully determined by a single random seed: the seed controls orbital parameter sampling, observation scheduling, noise injection, and $N$-body signal generation via \textsc{Rebound}~\citep{rein2012rebound} (details in Appendix~\ref{app:difficulty}).
Because every task is reproducible from its seed alone, the synthetic component is \emph{infinitely scalable}: new held-out suites can be generated on demand, preventing score saturation as models improve.
The 20 real-data tasks are constructed from published archival RV datasets; we provide conversion scripts that anonymise and reformat archival data into the \ourBenchmark{} interface, making it straightforward to incorporate additional real systems in the future (Appendix~\ref{app:real_data}).

\textbf{Difficulty scoring.}
Each task is assigned an integer difficulty $d \in [1,10]$ by summing six physics-based components derived from established RV theory and split into three difficulty levels~\citep{cumming2004detectability,angladaescude2010eccentric,queloz2001noplanet,haywood2014planets}: planet multiplicity, SNR, resonant configurations, period coverage, observation count, and correlated noise amplitude (Figure~\ref{fig:overview}, complete rubric in Appendix~\ref{app:difficulty}).
Factor weights were set \emph{a priori} from domain knowledge.
Tasks that were physically non-identifiable under the realized observation window and noise were filtered out to ensure solvability.

\subsection{Agentic Environment}

\begin{figure*}[t]
    \centering
    \includegraphics[width=1.0\textwidth]{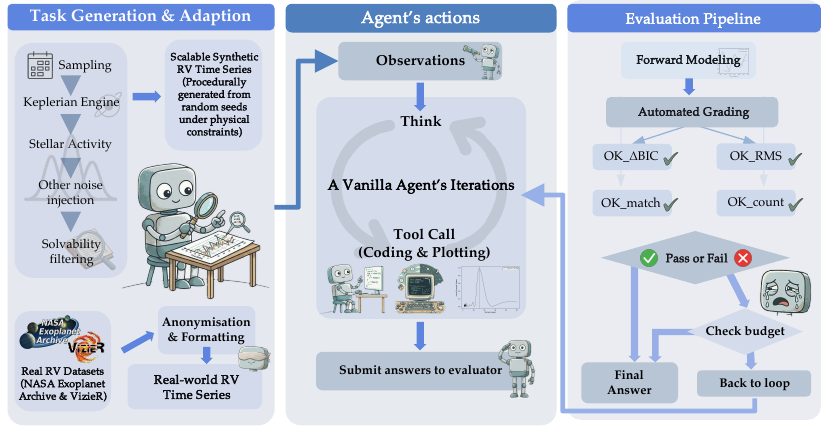}
    \vspace{-6pt}
    \caption{\ourBenchmark{} framework. \textbf{Left:} Task generation from synthetic physics or extracted from archival RV data. \textbf{Center:} Agent iteration loop of analysis, submission, and per-criterion feedback. \textbf{Right:} Evaluator forward-models submissions and grades with $\Delta$BIC, RMS, Match, and Count.}
    \label{fig:environment_overview}
\end{figure*}

As Figure~\ref{fig:environment_overview} illustrates, the environment provides the agent with the RV dataset (timestamps, velocities, uncertainties, and host star mass) at the beginning of each episode.
The agent operates in a ReAct-style loop with two tools: a \textbf{PythonREPL} for executing analysis code (periodograms, Keplerian fitting, residual inspection) and \textbf{submit\_action} interface for proposing candidate planetary systems. The agent may submit multiple times within an episode; only the best submission counts toward the final score. After each submission, the evaluator returns per-criterion diagnostic signals (pass/fail status and optional hints), enabling the agent to revise its hypothesis, for example by adding a planet if the count is wrong or refining parameters if the match score is low.

\textbf{Automated grading.}
The evaluator reconstructs the RV curve from the agent's submitted parameters via forward modeling, then applies four pass/fail criteria detailed in \S\ref{sec:evaluation}.

\textbf{Resource budgets.}
Each tier has a fixed resource budget calibrated from pilot runs: we observed the token and time consumption of successful trajectories and set limits at approximately $3\times$ the median successful cost to provide ample headroom while preventing runaway episodes.
The resulting budgets are 200K tokens / 600\,s (Easy), 450K / 900\,s (Medium), and 900K / 1500\,s (Hard), with 3, 5, and 10 submission attempts respectively.
An episode terminates when any limit is reached.

\subsection{Evaluation Protocol}
\label{sec:evaluation}

A task is considered solved only when \emph{all four} criteria below are simultaneously satisfied.

\textbf{Statistical metric.}
Given the observed radial velocities $\{y_i\}$, measurement uncertainties $\{\sigma_i\}$, and the agent's predicted model $\{\hat{y}_i\}$, the environment computes:

\textbf{(1) ok\_rms} (residual quality): the root-mean-square residual $\mathrm{RMS} = \sqrt{N^{-1}\sum_i(y_i - \hat{y}_i)^2}$ must satisfy $\mathrm{RMS} \le 1.5\,\tilde{\sigma}$, where $\tilde{\sigma}$ is the median reported measurement uncertainty. This ensures the model fits the data to within a factor of the observational noise floor.

\textbf{(2) ok\_delta\_bic} (model selection): the per-point $\Delta\mathrm{BIC}/N$ must be positive, where $\Delta\mathrm{BIC} = \mathrm{BIC}_{\mathrm{null}} - \mathrm{BIC}_{\mathrm{model}}$. The null model is a weighted-mean constant. The BIC is computed as $\mathrm{BIC} = -2\ln\mathcal{L} + k\ln N$, with $k = 5n_{\mathrm{pl}} + n_{\mathrm{inst}}$ free parameters (five Keplerian elements per planet plus one systemic velocity per instrument). A positive $\Delta\mathrm{BIC}/N$ confirms that the submitted model is statistically preferred over a flat line, penalising over-parameterised solutions.

\textbf{Physical metric.}
Submitted planets are matched to truth planets via the Hungarian algorithm~\citep{kuhn1955hungarian,budavari2016assignment,hopkins2015sourcefinding}
on a pairwise distance matrix:
\begin{equation}
  d_{ij} = 4.0\,\frac{\mathrm{RMS}(\mathrm{RV}_i - \mathrm{RV}_j)}{K_i}
         + 1.0\,\bigl|\ln\tfrac{P_j}{P_i}\bigr|
         + 0.5\,\bigl|\ln\tfrac{K_j}{K_i}\bigr|
         + 0.5\,|e_j - e_i|,
\end{equation}
where the RV-curve term compares single-planet Keplerian signals with an
optimal offset removed.  Its dominant weight ($w{=}4.0$) naturally absorbs
parameter degeneracies (e.g., the $\omega$/$M_0$ trade-off at low $e$).
Pairs with $d_{ij}>5$ are rejected.  The match score is
\begin{equation}
  S_{\mathrm{match}} = \frac{1}{|\mathcal{M}|}\sum_{(i,j)\in\mathcal{M}} e^{-d_{ij}}
    - 0.25\,\bigl|n_{\mathrm{truth}} - n_{\mathrm{guess}}\bigr|.
\end{equation}

\begin{figure}[t]
    \centering
    \includegraphics[width=\linewidth]{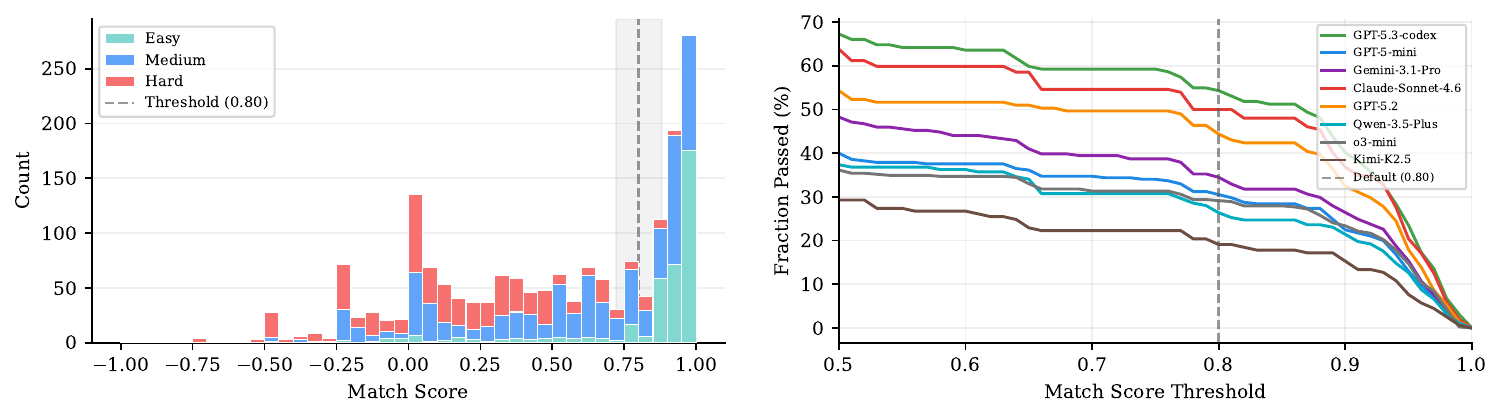}
    \vspace{-20pt}
    \caption{\textbf{(a)} Match-score distribution across all submitted episodes colored by difficulty tier. The shaded band marks the $\pm$10\% sensitivity region around the default threshold (0.80). \textbf{(b)} Pass rate as a function of the match threshold for each model. Rankings are preserved across the entire 0.5--1.0 range. Pass rates are computed as the unweighted fraction across all 100 synthetic tasks; per-tier results in Table~\ref{tab:main-results} remain our primary metric.}
    \label{fig:threshold_sensitivity}
\end{figure}

\textbf{(3) ok\_match}: $S_{\mathrm{match}} \ge 0.8$. The threshold $S_{\mathrm{match}} \ge 0.8$ corresponds to a mean pairwise distance $d \le 0.22$, requiring that submitted and ground-truth Keplerian signals closely overlap — a criterion more permissive than the parameter precisions typically reported in RV discovery papers (e.g., $\sigma_P/P < 1\%$, $\Delta e < 0.05$~\cite{Lovis2006}). The match-score distribution is strongly bimodal (Figure~\ref{fig:threshold_sensitivity}, left), with only 14\% of submissions in the $[0.72, 0.88]$ boundary region; sweeping the threshold by $\pm 10\%$ shifts pass rates by at most 5.0\,pp and preserves all model rankings (Figure~\ref{fig:threshold_sensitivity}, right).

\textbf{(4) ok\_count}: $n_{\mathrm{guess}} = n_{\mathrm{truth}}$.

The first two criteria verify that the submitted model fits the data; the latter two verify that it recovers the correct physical system. This conjunction gate prevents trivial solutions that achieve low residuals without identifying the right planets. All thresholds are fixed across tasks and models; only the RMS threshold adapts implicitly through $\tilde{\sigma}$, which varies with data quality. The match score is computed as the mean over successfully paired planets; unmatched planets are handled by the separate count-match criterion rather than included in the score itself. This conjunction gate captures orthogonal failure modes, as a model may pass \textbf{ok\_match} while failing \textbf{ok\_count} or vice versa, providing finer discrimination between models than either criterion alone.

\colorlet{stepcolor}{blue!60!black}
\colorlet{failcolor}{red!70!black}
\colorlet{successcolor}{green!50!black}
\colorlet{stepbg}{blue!10}
\colorlet{failbg}{red!10}
\colorlet{successbg}{green!15}
\raggedbottom

\section{Results and Discussion}
\label{sec:results}
\begin{table*}[t]
\centering
\resizebox{\textwidth}{!}{%
\begin{tabular}{l ccc ccc ccc c}
\toprule
& \multicolumn{3}{c}{\textbf{Pass Rate (\%)}}
& \multicolumn{3}{c}{\textbf{Env Done (\%)}}
& \multicolumn{3}{c}{\textbf{Pass@3 (\%)}}
& \textbf{Real} \\
\cmidrule(lr){2-4}\cmidrule(lr){5-7}\cmidrule(lr){8-10}\cmidrule(lr){11-11}
\textbf{Model}
  & \cellcolor{easy-bg}Easy & \cellcolor{med-bg}Med & \cellcolor{hard-bg}Hard
  & \cellcolor{easy-bg}Easy & \cellcolor{med-bg}Med & \cellcolor{hard-bg}Hard
  & \cellcolor{easy-bg}Easy & \cellcolor{med-bg}Med & \cellcolor{hard-bg}Hard
  & \cellcolor{real-bg}(20 tasks) \\
\midrule
\rowcolor{gray!10}
Classical Pipeline
  & \cellcolor{easy-bg}95.0 & \cellcolor{med-bg}35.0 & \cellcolor{hard-bg}5.0
  & \cellcolor{easy-bg}100 & \cellcolor{med-bg}100 & \cellcolor{hard-bg}100
  & \cellcolor{easy-bg}95.0 & \cellcolor{med-bg}35.0 & \cellcolor{hard-bg}5.0
  & \cellcolor{real-bg}---\\
\rowcolor{gray!10}
Nested Sampling
  & \cellcolor{easy-bg}95.0 & \cellcolor{med-bg}32.5 & \cellcolor{hard-bg}0.0
  & \cellcolor{easy-bg}100 & \cellcolor{med-bg}100 & \cellcolor{hard-bg}100
  & \cellcolor{easy-bg}95.0 & \cellcolor{med-bg}32.5 & \cellcolor{hard-bg}0.0
  & \cellcolor{real-bg}---\\
\midrule
o3-mini
  & \cellcolor{easy-bg}40.0 & \cellcolor{med-bg}24.2 & \cellcolor{hard-bg}0.0
  & \cellcolor{easy-bg}\second{76.7} & \cellcolor{med-bg}35.8 & \cellcolor{hard-bg}4.2
  & \cellcolor{easy-bg}\best{95.0} & \cellcolor{med-bg}\best{40.0} & \cellcolor{hard-bg}0.0
  & \cellcolor{real-bg}\textcolor{red}{0.0}\\
GPT-5-mini
  & \cellcolor{easy-bg}\second{76.7} & \cellcolor{med-bg}\second{33.6} & \cellcolor{hard-bg}2.5
  & \cellcolor{easy-bg}\second{76.7} & \cellcolor{med-bg}\second{46.7} & \cellcolor{hard-bg}5.0
  & \cellcolor{easy-bg}\best{95.0} & \cellcolor{med-bg}35.0 & \cellcolor{hard-bg}2.5
  & \cellcolor{real-bg}\textcolor{red}{0.0}\\
GPT-5.2
  & \cellcolor{easy-bg}40.0 & \cellcolor{med-bg}{30.0} & \cellcolor{hard-bg}\best{5.8}
  & \cellcolor{easy-bg}40.0 & \cellcolor{med-bg}33.3 & \cellcolor{hard-bg}\second{5.8}
  & \cellcolor{easy-bg}75.0 & \cellcolor{med-bg}\second{37.5} & \cellcolor{hard-bg}\best{12.5}
  & \cellcolor{real-bg}\textcolor{red}{0.0}\\
Kimi-K2.5
  & \cellcolor{easy-bg}13.3 & \cellcolor{med-bg}17.9 & \cellcolor{hard-bg}0.8
  & \cellcolor{easy-bg}13.3 & \cellcolor{med-bg}19.2 & \cellcolor{hard-bg}2.5
  & \cellcolor{easy-bg}40.0 & \cellcolor{med-bg}35.0 & \cellcolor{hard-bg}2.5
  & \cellcolor{real-bg}\textcolor{red}{0.0}\\
Qwen-3.5-Plus
  & \cellcolor{easy-bg}26.7 & \cellcolor{med-bg}25.0 & \cellcolor{hard-bg}1.6
  & \cellcolor{easy-bg}25.0 & \cellcolor{med-bg}25.8 & \cellcolor{hard-bg}1.7
  & \cellcolor{easy-bg}60.0 & \cellcolor{med-bg}30.0 & \cellcolor{hard-bg}2.5
  & \cellcolor{real-bg}\textcolor{red}{0.0}\\
Gemini-3.1-Pro
  & \cellcolor{easy-bg}71.7 & \cellcolor{med-bg}\best{35.0} & \cellcolor{hard-bg}\second{5.0}
  & \cellcolor{easy-bg}71.7 & \cellcolor{med-bg}35.0 & \cellcolor{hard-bg}\second{5.8}
  & \cellcolor{easy-bg}\best{95.0} & \cellcolor{med-bg}35.0 & \cellcolor{hard-bg}\second{7.5}
  & \cellcolor{real-bg}\textcolor{red}{0.0}\\
Claude-Sonnet-4.6
  & \cellcolor{easy-bg}68.3 & \cellcolor{med-bg}22.5 & \cellcolor{hard-bg}0.8
  & \cellcolor{easy-bg}68.3 & \cellcolor{med-bg}28.3 & \cellcolor{hard-bg}0.8
  & \cellcolor{easy-bg}75.0 & \cellcolor{med-bg}32.5 & \cellcolor{hard-bg}2.5
  & \cellcolor{real-bg}\textcolor{red}{0.0}\\
GPT-5.3-codex
  & \cellcolor{easy-bg}\best{80.0} & \cellcolor{med-bg}{30.8} & \cellcolor{hard-bg}{4.2}
  & \cellcolor{easy-bg}\best{88.3} & \cellcolor{med-bg}\best{48.3} & \cellcolor{hard-bg}\best{7.5}
  & \cellcolor{easy-bg}\best{95.0} & \cellcolor{med-bg}\best{40.0} & \cellcolor{hard-bg}\second{7.5}
  & \cellcolor{real-bg}\textcolor{red}{0.0}\\
\bottomrule
\end{tabular}%
}
\caption{Main results on \ourBenchmark{} across difficulty tiers computed over three independent runs. Pass rates and Env Done rates are averaged; Pass@3 reports the fraction of tasks solved in at least one run. \textbf{Bold} = best per column; \underline{underline} = second best.}
\label{tab:main-results}
\vspace{-6pt}
\end{table*}

\textbf{Baselines without LLMs.}
We include two baselines using deterministic programs without LLMs.
The \textbf{Classical Pipeline} chains Lomb-Scargle periodogram search, weighted circular-orbit initialization, multi-start Keplerian fitting (\texttt{scipy.optimize.least\_squares}), and greedy BIC-gated planet addition ($\Delta\mathrm{BIC} > 10$).
The \textbf{Nested Sampling} baseline uses Bayesian model comparison via nested sampling to select the number of planets, then fits orbital parameters under the best model.
Both achieve 95.0\% on Easy (stronger than the best LLM agents) but degrade on harder tiers (Classical: 35.0\% Medium, 5.0\% Hard; Nested Sampling: 32.5\% Medium, 0.0\% Hard).
They shared inability to reliably detect more than one planet (average predicted count $\approx$1.1 across all tiers).
These baselines demonstrate LLM agents have not yet outperformed traditional methods on simple single-planet tasks.

\textbf{Performance drops sharply as difficulty increases.}
As Table~\ref{tab:main-results} shows, top performers exceed 70\% on Easy tasks, but drops to 17--35\% on Medium and collapses on Hard, where no model exceeds 6\%.
This degradation is consistent across all models, which confirms the construct validity of our physics-grounded stratification.
GPT-5.3-codex achieves the best Easy-tier pass rate (80.0\%) and the highest natural completion rate across all tiers, indicating that it reaches termination conditions within budget more often than other models.
In contrast, o3-mini completes 76.7\% of Easy tasks but has a 0\% pass rate on Hard, where it often stop early without producing correct results.
We also found that the dominant difficulty drivers correlated with lower success rates are SNR and planet multiplicity measured by Pearson Correlation (Appendix~\ref{app:difficulty_factor_correlations}).

\textbf{Frontier agents fail consistently on real-data subset.}
The \textbf{Real} column of Table~\ref{tab:main-results} reports performance on the 20 tasks based on real-world RV data (Appendix~\ref{app:real_data}).
Across three runs, no model achieves a single pass (0\% for all 8 models).
Among episodes that produce a submission, $\Delta$BIC passes universally (100\%) and RMS passes in 40\%, but Match Score remains at 0\% and Count passes in only 27\%: the closest cases recover correct orbital periods but overestimate semi-amplitudes (Table~\ref{tab:real_data_near_matches}), failing to fully characterise the underlying planets.

All real-data tasks have been solved by human astronomers as the ground-truth are taken from peer-reviewed papers with confirmed, published solutions.
As an independent validation step, we re-fit each system's RV data using \texttt{RadVel}~\citep{fulton2018radvel} and verified that the recovered parameters agree with the published values within their reported uncertainties (Appendix~\ref{app:real_data}).
These tasks are therefore \emph{provably solved} by human experts under the same condition, reflecting a critical capability gap frontier agents have yet to reach human researcher level.
This also provides evidence against data contamination: although these published papers could have appeared in training corpora, no frontier models were able to achieve a single success on this subset.

\subsection{Statistical Fitting vs.\ Physics Reasoning}

\begin{figure}[t]
    \centering
    \includegraphics[width=\linewidth]{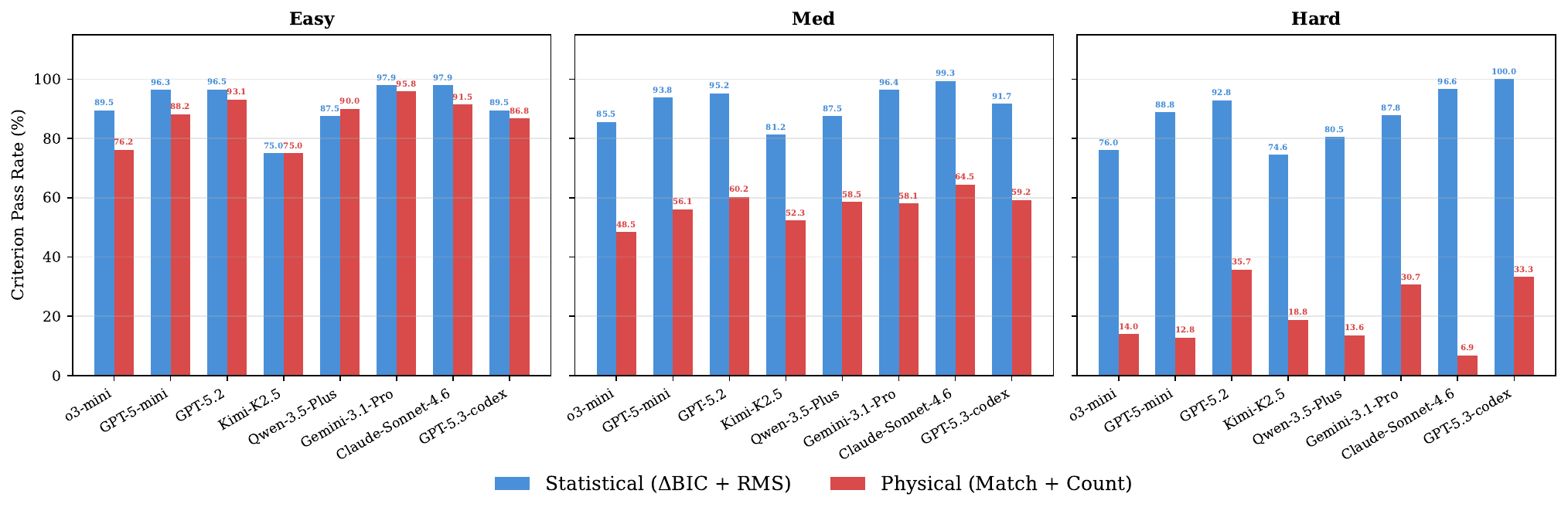}
    \vspace{-20pt}
    \caption{Statistical (blue, mean of $\Delta$BIC and RMS) versus physical (red, mean of Match Score and Planet Count) criterion pass rates by difficulty tier. Statistical pass rates stay high while physical recovery drops from Easy to Hard.}
    \label{fig:stat_vs_phys}
\end{figure}

Figure~\ref{fig:stat_vs_phys} decomposes pass rates into the four evaluation criteria.
Statistical criteria ($\Delta$BIC and RMS) remain above 70\% across all models and tiers, confirming that agents reliably produce well-fitting models.
Physical criteria (Match Score and Planet Count), by contrast, drop below 40\% on Hard tasks for every model.
This gap implies that models optimize within a fixed hypothesis (curve fitting) rather than searching over physically plausible configurations (model selection), which is the core reasoning bottleneck that \ourBenchmark{} is designed to expose.

\subsection{Test-Time Scaling, Resource Budget and Completion Rate}
\label{app:json_repair}

\textbf{Test-Time Scaling in Pass@3}
We found pass@3 based on 3 independent runs lead to substantial stochasticity: four models reach 95\% Pass@3 on Easy-tier despite much lower mean pass rates of 40--80\%, and GPT-5.2 attains the highest Hard-tier Pass@3 (12.5\%). It's worth noting that the 3 runs are fully independent and do not leak any information/experience from 1 run to another, the improvement here is simply a result of more attempts rather than longer reasoning.

\textbf{Test-Time Scaling in Single Run.} On the other hand, we have a reasoning token budget for each independent run and report how often agents finish within the budget constraints (token limit, timeout, or tool-call cap) in the Env Done columns in Table~\ref{tab:main-results}.
In hard tasks, six of eight models complete fewer than 6\% of tasks naturally, where the majority is cut off by budget limits.
GPT-5.3-codex is a notable outlier, completing 7.5\% of Hard tasks naturally, driven by its compact output style ($\sim$4K tokens per task).

\subsection{Self-Generated Skills.}\label{sec:skills}

Following~\citet{li2025skillsbench}, we extract a \texttt{skills.md} summary from successful Easy-tier trajectories generated by Opus 4.6 (which is \emph{not} among the eight models evaluated in the main experiment, so the skills document is fully independent of the evaluation results).
The skill is then provided to 4 tested models at inference time (Table~\ref{tab:skills_results}), which shows boost to Easy-tier pass tasks for three of four models (up to +28.3\,pp).
\begin{table*}[t]
  \centering
  \resizebox{\textwidth}{!}{%
  \setlength{\tabcolsep}{4pt}
  \footnotesize
  \begin{tabular}{l ccc ccc ccc ccc}
  \toprule
  & \multicolumn{6}{c}{\textbf{Pass Rate (\%)}}
  & \multicolumn{6}{c}{\textbf{Pass@3 (\%)}} \\
  \cmidrule(lr){2-7}\cmidrule(lr){8-13}
  & \multicolumn{3}{c}{Default} & \multicolumn{3}{c}{+\,Skills}
  & \multicolumn{3}{c}{Default} & \multicolumn{3}{c}{+\,Skills} \\
  \cmidrule(lr){2-4}\cmidrule(lr){5-7}\cmidrule(lr){8-10}\cmidrule(lr){11-13}
  \textbf{Model}
    & \cellcolor{easy-bg}Easy & \cellcolor{med-bg}Med & \cellcolor{hard-bg}Hard
    & \cellcolor{skill-easy-bg}Easy & \cellcolor{skill-med-bg}Med & \cellcolor{skill-hard-bg}Hard
    & \cellcolor{easy-bg}Easy & \cellcolor{med-bg}Med & \cellcolor{hard-bg}Hard
    & \cellcolor{skill-easy-bg}Easy & \cellcolor{skill-med-bg}Med & \cellcolor{skill-hard-bg}Hard \\
  \midrule
  GPT-5-mini
    & \cellcolor{easy-bg}76.7 & \cellcolor{med-bg}33.6 & \cellcolor{hard-bg}2.5
    & \cellcolor{skill-easy-bg}90.0\textsubscript{\textcolor{ForestGreen}{\tiny+13.3}}
    & \cellcolor{skill-med-bg}35.8\textsubscript{\textcolor{ForestGreen}{\tiny+2.2}}
    & \cellcolor{skill-hard-bg}2.5\textsubscript{\textcolor{gray}{\tiny+0.0}}
    & \cellcolor{easy-bg}95.0 & \cellcolor{med-bg}35.0 & \cellcolor{hard-bg}2.5
    & \cellcolor{skill-easy-bg}95.0\textsubscript{\textcolor{gray}{\tiny+0.0}}
    & \cellcolor{skill-med-bg}37.5\textsubscript{\textcolor{ForestGreen}{\tiny+2.5}}
    & \cellcolor{skill-hard-bg}2.5\textsubscript{\textcolor{gray}{\tiny+0.0}} \\
  Qwen-3.5-Plus
    & \cellcolor{easy-bg}26.7 & \cellcolor{med-bg}25.0 & \cellcolor{hard-bg}1.6
    & \cellcolor{skill-easy-bg}48.3\textsubscript{\textcolor{ForestGreen}{\tiny+21.6}}
    & \cellcolor{skill-med-bg}23.3\textsubscript{\textcolor{red}{\tiny$-$1.7}}
    & \cellcolor{skill-hard-bg}0.0\textsubscript{\textcolor{red}{\tiny$-$1.6}}
    & \cellcolor{easy-bg}60.0 & \cellcolor{med-bg}30.0 & \cellcolor{hard-bg}2.5
    & \cellcolor{skill-easy-bg}90.0\textsubscript{\textcolor{ForestGreen}{\tiny+30.0}}
    & \cellcolor{skill-med-bg}35.0\textsubscript{\textcolor{ForestGreen}{\tiny+5.0}}
    & \cellcolor{skill-hard-bg}0.0\textsubscript{\textcolor{red}{\tiny$-$2.5}} \\
  Gemini-3.1-Pro
    & \cellcolor{easy-bg}71.7 & \cellcolor{med-bg}35.0 & \cellcolor{hard-bg}5.0
    & \cellcolor{skill-easy-bg}100.0\textsubscript{\textcolor{ForestGreen}{\tiny+28.3}}
    & \cellcolor{skill-med-bg}56.7\textsubscript{\textcolor{ForestGreen}{\tiny+21.7}}
    & \cellcolor{skill-hard-bg}16.7\textsubscript{\textcolor{ForestGreen}{\tiny+11.7}}
    & \cellcolor{easy-bg}95.0 & \cellcolor{med-bg}35.0 & \cellcolor{hard-bg}7.5
    & \cellcolor{skill-easy-bg}100.0\textsubscript{\textcolor{ForestGreen}{\tiny+5.0}}
    & \cellcolor{skill-med-bg}80.0\textsubscript{\textcolor{ForestGreen}{\tiny+45.0}}
    & \cellcolor{skill-hard-bg}35.0\textsubscript{\textcolor{ForestGreen}{\tiny+27.5}} \\
  GPT-5.3-codex
    & \cellcolor{easy-bg}80.0 & \cellcolor{med-bg}30.8 & \cellcolor{hard-bg}4.2
    & \cellcolor{skill-easy-bg}74.6\textsubscript{\textcolor{red}{\tiny$-$5.4}}
    & \cellcolor{skill-med-bg}53.2\textsubscript{\textcolor{ForestGreen}{\tiny+22.4}}
    & \cellcolor{skill-hard-bg}25.6\textsubscript{\textcolor{ForestGreen}{\tiny+21.4}}
    & \cellcolor{easy-bg}95.0 & \cellcolor{med-bg}40.0 & \cellcolor{hard-bg}7.5
    & \cellcolor{skill-easy-bg}100.0\textsubscript{\textcolor{ForestGreen}{\tiny+5.0}}
    & \cellcolor{skill-med-bg}67.5\textsubscript{\textcolor{ForestGreen}{\tiny+27.5}}
    & \cellcolor{skill-hard-bg}31.6\textsubscript{\textcolor{ForestGreen}{\tiny+24.1}} \\
  \bottomrule
  \end{tabular}%
  }
    \caption{Effect of domain-expert skills injection on pass rates (\%) and Pass@3 (\%), averaged over three independent runs.
  Colored superscripts show the change in percentage points
  (\textcolor{ForestGreen}{$+$} = improvement, \textcolor{red}{$-$} = degradation).}
  \label{tab:skills_results}
  \vspace{-6pt}
  \end{table*}
However, an episode-level analysis (Appendix~\ref{app:skills_stop_reasons},~\ref{app:skills_criteria}) reveals that these gains are largely driven by \emph{efficiency} rather than \emph{reasoning}: skills compress the workflow so that previously budget-exceeding episodes now reach the submission stage, although Hard-tier Match Score remains below 33\% for all models.
For weaker models, skills even degrade Hard-tier RMS, suggesting that rigid templates may have interfered with the more advanced strategies needed for complex systems.
The statistical-physical dissociation from Section~4.2 persists across all models.

  \subsection{Case Studies}
  \label{sec:case_studies}

  We trace two representative trajectories---one success and one failure---that illustrate the behavioral divide between agents that escalate model complexity and those
  that do not (full step-by-step traces in Appendix~\ref{app:case_studies}).

  \begin{tcolorbox}[
    enhanced, breakable,
    colback=white, colframe=soft-line,
    colbacktitle=pantone655, coltitle=white,
    boxrule=0.6pt, arc=1.5pt,
    fonttitle=\bfseries\footnotesize,
    toptitle=1pt, bottomtitle=1pt,
    left=3pt, right=3pt, top=2pt, bottom=2pt,
    title={\cmark\;\;Success Case: 2-Planet Recovery
      \hfill \normalfont\scriptsize\itshape GPT-5.2\,$|$\,11 steps}
  ]

  {\footnotesize
  \begin{tabularx}{\linewidth}{@{}lXlX@{}}
    \toprule
    \multicolumn{4}{@{}l}{\textbf{Task}\enspace Seed \texttt{96} (Medium)} \\
    \midrule
    \textbf{Planets}       & 2\; ($P = 31.2,\;117.6$\,d) &
    \textbf{Observations}  & 75 \\
    \textbf{Noise model}   & White\; ($\sigma_w = 1.05$\,m\,s$^{-1}$) &
    \textbf{SNR}           & 8.9 \\
    \midrule
    \multicolumn{4}{@{}p{\linewidth}@{}}{%
      \textbf{Difficulty.}\;
      Planet~2 dominates the RV signal ($K_2 \approx 30$\,m\,s$^{-1}$, $e_2 = 0.25$), but its period ($117.6$\,d) exceeds the data span ($106.7$\,d), causing the
  periodogram to converge on a biased estimate ($\sim 105$\,d).
      Planet~1 ($P = 31.2$\,d, $K_1 \approx 14$\,m\,s$^{-1}$) is buried in the residuals.} \\
    \bottomrule
  \end{tabularx}
  }
  {\footnotesize\raggedright
  \medskip
  \noindent
  \colorbox{stepbg}{\textbf{Steps 1--5 · Protocol \& Baseline}}\quad
  Reads protocol, computes weighted Lomb--Scargle periodogram (40\,000 frequencies).
  Dominant peak at $P \approx 105$\,d (power $= 0.80$).
  Baseline one-sine: $P = 104.8$\,d, RMS $= 10.68$\,m\,s$^{-1}$ (RMS$/\tilde\sigma = 10.1$).
  Gate: \textbf{Kepler = YES.}\par\vspace{3pt}

  \noindent
  \colorbox{stepbg}{\textbf{Steps 6--9 · Iterative Keplerian Fitting}}\quad
  1-planet Keplerian converges to $P = 300$\,d at the search boundary (RMS $= 9.16$\,m\,s$^{-1}$)---a diagnostic sign that the true period exceeds the data span.
  Residual periodogram reveals a strong second signal at $P \approx 31.0$\,d (power $= 0.84$).
  2-planet fit: $P_1 = 116.6$\,d, $P_2 = 31.1$\,d; RMS drops to $\mathbf{1.13}$\,m\,s$^{-1}$ (RMS$/\tilde\sigma = 1.07$).\par\vspace{3pt}

  \noindent
  \colorbox{successbg}{\textbf{Step 11 · Submit \#1 (2 planets) $\to$ Pass}}\quad
  Match $= 0.932$,\ RMS $= 1.13$\,m\,s$^{-1}$,\ $\Delta$BIC/pt $= 513$.\par
  }
  \diagpass{bic}\;\diagpass{rms}\;\diagpass{match}\;\diagpass{count}\\[3pt]
  {\small\textbf{Resources:}\; 61\,K input + 7\,K output tokens, 0 code errors, 1 submission.}
  \begin{insightbox}
  The agent follows a textbook \emph{peel-and-search} strategy: fit the strongest
  signal, subtract, search residuals for the next.
  A key diagnostic moment occurs when the 1-planet fit converges to the period
  upper bound ($P = 300$\,d): instead of submitting, the agent recognises
  incomplete convergence and checks residuals, where the 31\,d signal guides
  the 2-planet fit that simultaneously corrects $P_1$ to $116.6$\,d.
  Both planets are recovered in 11 steps with a single submission, consuming
  only 68\,K tokens.
  \end{insightbox}
  \end{tcolorbox}

  \noindent Successful agents treat failed submissions as informative feedback: they use mismatched diagnostics (e.g., low Match despite good RMS) to revise the hypothesis, escalate model complexity (adding planets), and validate the new fit via residual checks and follow-up periodograms before resubmitting. In contrast, failed agents often lock onto a single hypothesis and fall into repetitive resubmissions that do not incorporate new evidence, exhausting the step and token budget without meaningful exploration. 
  \begin{tcolorbox}[
    enhanced, breakable,
    colback=white, colframe=soft-line,
    colbacktitle=pantone655, coltitle=white,
    boxrule=0.6pt, arc=1.5pt,
    fonttitle=\bfseries\footnotesize,
    toptitle=1pt, bottomtitle=1pt,
    left=3pt, right=3pt, top=2pt, bottom=2pt,
    title={\xmark\;\;Failure Case: Stuck in a Local Minimum
      \hfill \normalfont\scriptsize\itshape GPT-5-mini\,$|$\,40 steps}
  ]

  {\footnotesize
  \begin{tabularx}{\linewidth}{@{}lXlX@{}}
    \toprule
    \multicolumn{4}{@{}l}{\textbf{Task}\enspace Seed \texttt{196} (Hard)} \\
    \midrule
    \textbf{Planets}       & 3\; ($P = 76,\;112,\;167$\,d) &
    \textbf{Observations}  & 43 \\
    \textbf{Noise model}   & White\; ($\sigma_w = 0.93$\,m\,s$^{-1}$) &
    \textbf{Resonance}     & Double\; ($\approx 3{:}2{:}1$) \\
    \midrule
    \multicolumn{4}{@{}p{\linewidth}@{}}{%
      \textbf{Difficulty.}\;
      The three planets sit in a double-resonance chain with period ratios near
      $3{:}2{:}1$, producing alias peaks and combination frequencies in the
      periodogram that make it easy for an iterative search to lock onto
      spurious solutions instead of the true orbital periods.} \\
    \bottomrule
  \end{tabularx}
  }
  {\footnotesize\raggedright
  \medskip
  \noindent
  \colorbox{stepbg}{\textbf{Steps 1--12 · Protocol, Periodogram \& Fitting}}\quad
  Runs the standard pipeline: protocol, periodogram, baseline, and iterative fitting.
  Converges on a 2-planet model with alias periods $(111.7, 164.5)$\,d that achieve
  low RMS ($1.08$\,m\,s$^{-1}$) but do not correspond to real planets.\\[3pt]
  \colorbox{failbg}{\textbf{Step 13 · Submit \#1 (2 planets) $\to$ Fail}}\quad
  RMS $= 7.71$\,m\,s$^{-1}$; recovered parameters are wildly incorrect due to
  \texttt{l\_rad} format error.\\[3pt]
  \colorbox{failbg}{\textbf{Steps 14--40 · 9 More Submissions $\to$ All Fail}}\quad
  By step~20 the format error is fixed and RMS $= 1.08$\,m\,s$^{-1}$,
  but match $= -0.008$: the detected periods are aliases of the true resonant system.
  \textbf{Submissions 3--10 are near-identical}: the same wrong
  answer is resubmitted without any model escalation.\\[3pt]}
  {\small\textbf{Resources:}\; 696\,K input + 34\,K output tokens, 14 code errors, 10 submissions, all failed.}\\[2pt]
  {\small\textbf{Diagnostics:}\;
  \diagpass{bic}\;\diagpass{rms}\;\diagfail{match}\;\diagfail{count (2 vs.\ 3 true)}}

  \begin{insightbox}
  The agent converges to a 2-planet model that achieves good RMS but corresponds
  to \emph{alias periods}, a classic RV pitfall where near-resonant planets produce
  periodogram peaks at combination frequencies. After reaching this local minimum,
  the agent cannot escape: it repeats the same submission many times without
  attempting a 3-planet model.
  The failure consumes $10.7\times$ more tokens (730\,K vs.\ 68\,K) than the
  success case, demonstrating that \emph{more computation does not compensate
  for the inability to revise the model hypothesis}.
  \end{insightbox}

  \end{tcolorbox}

\subsection{Takeaway for Future Model Development}

The evaluation of frontier agents within \ourBenchmark{} offers several insights for next-generation AI scientists. More broadly, \ourBenchmark{} also provides a scalable environment for agentic RL in iterative, feedback-driven scientific workflows:

\textbf{Goodness-of-fit $\neq$ physical recovery.}
Future model developers should pay more attention to the construct validity of their evaluation metric.
High scores on statistical goodness-of-fit do not necessarily translate to recovering meaningful physical parameters.
Training agents for science should move beyond optimizing for residuals and incorporate metrics that explicitly test scientific validity in the context of the task.
More broadly, agent policies should treat diagnostic mismatches as triggers for hypothesis revision and model-complexity escalation, rather than as signals to spend more compute on the same solution.

\textbf{Procedural scaffolding has limits.}
On relatively simple tasks, self-generated skills improve efficiency of agentic performance, but they do not fundamentally improve physical reasoning on Hard tasks.
This finding is consistent with recent studies on the limitation of skills~\cite{li2025skillsbench}, where domain expertise is required to curate genuinely generalizable skills.
In practice, it may be beneficial to pair procedural scaffolds with robustness mechanisms such as strict output-format compliance and automated harness.

\textbf{Mind the sim-to-real gap.}
For agentic RL training, the sim-to-real gap should be treated as a primary consideration. It can determine whether policies learned under dense and well-structured simulated feedback transfer to real RV data that exhibit noise, sparse sampling, and instrument systematics.
We therefore suggest evaluating transfer explicitly and designing physics-grounded data mixtures and curricula that disentangle what is controllable in simulation from what must be handled by the agent at deployment time.

\section{Conclusion}
We introduced \ourBenchmark{}, a scalable environment for AI agents on the iterative, multi-step workflow of exoplanet discovery via radial-velocity analysis. By moving beyond static QA toward dynamic, feedback-driven scientific reasoning, \ourBenchmark{} exposes capability gaps that existing benchmarks cannot detect. Our evaluation of eight frontier models reveals a consistent pattern: agents are proficient at numerical optimization but struggle with the physical reasoning that distinguishes curve fitting from scientific discovery. This statistical--physical dissociation persists across models, difficulty tiers, and even when agents are equipped with domain-expert skills or additional compute budget. By quantifying these limitations in a physically grounded, infinitely scalable setting, \ourBenchmark{} provides a concrete target for developing more capable scientific agents that can not only fit data, but also interpret what the fit means.

\section*{Acknowledgements}

This work is supported in part by the National Science and Engineering Research Council of Canada, the Dunlap Institute of Astronomy \& Astrophysics (seed funding), Anthropic (compute credits), OpenAI (superalignment grant), the German Federal Ministry of Education and Research (BMBF) and T\"ubingen AI Center (FKZ: 01IS18039B), and the Machine Learning Cluster of Excellence (EXC 2064/1, Project 390727645).

\clearpage
\section*{LLM Usage Statement}
We used large language models to assist with drafting and revising prose and with minor \LaTeX{} editing. All technical content, experimental design, code, analyses, and results were produced and verified by the authors.

\bibliographystyle{colm2026_conference}
\bibliography{main}

\clearpage
\appendix
\startcontents[app]
\printcontents[app]{l}{1}{\section*{Appendix Table of Contents}}

\section{Task Construction Details}
\label{app:task_construction}

\subsection{Difficulty Scoring Rubric}
\label{app:difficulty}

Each synthetic task is assigned an integer difficulty level $d \in [1, 10]$ based on six physically motivated factors.
The difficulty score is computed as

\begin{equation}
d = \mathrm{clip}\!\left(d_{\mathrm{base}} + d_{\mathrm{SNR}} + d_{\mathrm{res}} + d_{\mathrm{cov}} + d_{\mathrm{obs}} + d_{\mathrm{GP}},\; 1,\; 10\right),
\end{equation}

\noindent where each component is defined in Table~\ref{tab:difficulty_factors}.

\begin{table}[h!]
\centering
\small
\begin{tabular}{lll}
\toprule
\textbf{Factor} & \textbf{Condition} & \textbf{Score} \\
\midrule
\multirow{4}{*}{Planet multiplicity ($d_{\mathrm{base}}$)}
  & 1 planet  & +1 \\
  & 2 planets & +2 \\
  & 3 planets & +3 \\
  & 4+ planets & +4 \\
\midrule
\multirow{4}{*}{Signal-to-noise ratio ($d_{\mathrm{SNR}}$)}
  & SNR $> 5$   & 0 \\
  & SNR $> 2$   & +1 \\
  & SNR $> 1$   & +2 \\
  & SNR $\leq 1$ & +3 \\
\midrule
\multirow{2}{*}{Resonant configurations ($d_{\mathrm{res}}$)}
  & 0 resonances & 0 \\
  & $\geq$1 resonance & $+\min(2,\, n_{\mathrm{res}})$ \\
\midrule
\multirow{3}{*}{Coverage of inner period ($d_{\mathrm{cov}}$)}
  & $T_{\mathrm{base}} / P_{\mathrm{inner}} \geq 3$ & 0 \\
  & $\geq 2$ & +1 \\
  & $< 2$    & +2 \\
\midrule
\multirow{4}{*}{Number of observations ($d_{\mathrm{obs}}$)}
  & $n_{\mathrm{obs}} \geq 80$ & 0 \\
  & $\geq 50$ & +1 \\
  & $\geq 30$ & +2 \\
  & $< 30$    & +3 \\
\midrule
\multirow{4}{*}{Correlated noise ($d_{\mathrm{GP}}$)}
  & No GP       & 0 \\
  & $\sigma_{\mathrm{GP}} < 0.5$ & +1 \\
  & $\sigma_{\mathrm{GP}} < 1.0$ & +2 \\
  & $\sigma_{\mathrm{GP}} \geq 1.0$ & +3 \\
\bottomrule
\end{tabular}
\caption{Difficulty scoring rubric. Each factor contributes an additive term to the total difficulty score, which is clipped to $[1, 10]$.}
\label{tab:difficulty_factors}
\end{table}

\subsection{Synthetic Task Generation Details}
\label{app:generation_details}

Each synthetic task is fully determined by a single random seed. The generation pipeline draws parameters from the following priors:

\begin{itemize}
\item \textbf{Number of planets:} 1--4, with higher counts at higher difficulty levels.
\item \textbf{Orbital periods:} log-uniform from $[2, 300]$\,days, with a 25\% probability of inserting near-resonant pairs (period ratios within 3\% of 2:1, 3:2, or 5:3).
\item \textbf{Minimum masses:} $m\sin i$ drawn from $[0.01, 1.0]\,M_\mathrm{Jup}$.
\item \textbf{Eccentricities:} Kipping Beta distribution~\citep{kipping2013parametrizing} with $\alpha=0.867$, $\beta=3.03$.
\item \textbf{Angular parameters:} argument of periastron $\omega$, longitude of ascending node $\Omega$, and mean longitude $\ell$ each drawn uniformly from $[0, 2\pi)$.
\item \textbf{White noise:} measurement uncertainty $\sigma_w$ drawn from $10^{U(-0.3,\,0.7)}$\,m\,s$^{-1}$ (approximately 0.5--5\,m\,s$^{-1}$), with an optional jitter term $\sigma_j$ added in quadrature.
\item \textbf{Correlated stellar noise:} included with 40\% probability. Modeled as a Gaussian Process with a quasi-periodic rotation kernel (celerite2 \texttt{RotationTerm}), parameterized by amplitude $\sigma_\mathrm{GP} \in [0.05, 1.6]$\,m\,s$^{-1}$ and stellar rotation period $\in [10, 45]$\,days.
\item \textbf{Observation schedule:} timestamps drawn uniformly over a baseline spanning 2--4$\times$ the shortest planetary period, producing an irregularly sampled time grid typical of ground-based surveys. The number of observations ranges from 30 to 100.
\end{itemize}

The clean RV signal is computed as a multi-Keplerian superposition via numerical solution of Kepler's equation (Newton iteration). Per-instrument systemic velocity offsets $\gamma_i$ are added for multi-instrument tasks.

Tasks are grouped into three tiers: \textbf{Easy} (difficulty 1--2, 20 tasks), \textbf{Medium} (difficulty 3--6, 40 tasks), and \textbf{Hard} (difficulty 7--10, 40 tasks).
Easy tasks are intentionally fewer because low-complexity single-planet systems occupy a smaller region of the physically plausible parameter space.
The scoring rubric was calibrated through pilot experiments in two steps.
We first assigned provisional weights by reverse-engineering which physical factors most reliably destabilise the standard RV workflow used by human analysts: low SNR, multiplicity, resonances, poor coverage, limited observations, and correlated stellar noise.
We then adjusted the weights and tier cutoffs so that pilot pass rates decreased monotonically with nominal difficulty, while keeping the score interpretable as a sum of physically meaningful failure drivers rather than a purely empirical hardness label.
Candidate instances that were physically non-identifiable under the realised cadence and noise draw were removed during benchmark construction, so difficulty is calibrated over a pool of tasks intended to be challenging but still solvable in principle.

\subsection{Real-World RV Dataset}
\label{app:real_data}

In addition to the 100 synthetic tasks, \textsc{Stargazer} includes 20 tasks constructed from published radial velocity datasets of confirmed exoplanetary systems.
These tasks span the full range of RV analysis complexity: from single hot Jupiters with $K > 100$\,m\,s$^{-1}$ to multi-planet systems with sub-m\,s$^{-1}$ signals buried in correlated noise.
All identifying information (target names, instrument names, literature references, prior knowledge of planetary parameters) is removed from the task files presented to the agent; the agent receives only time-series data, measurement uncertainties, instrument labels (anonymised as \texttt{inst\_A}, \texttt{inst\_B}, \ldots), and the host star mass.

Table~\ref{tab:real_data_tasks} lists the 20 real-data tasks with their provenance.
The systems are ordered by estimated analysis difficulty, which reflects the number of planets, signal-to-noise ratio, orbital architecture (resonances, high eccentricity), and data complexity (number of instruments, stellar activity).

\begin{table*}[!htbp]
\centering
\small
\resizebox{\textwidth}{!}{%
\begin{tabular}{cl cccc ll}
\toprule
\textbf{ID} & \textbf{System} & $N_\mathrm{obs}$ & $N_\mathrm{pl}$ & $K_\mathrm{max}$ (m\,s$^{-1}$) & $N_\mathrm{inst}$ & \textbf{Key Challenge} & \textbf{Reference} \\
\midrule
\multicolumn{8}{l}{\textit{Single-planet systems}} \\
\addlinespace
real\_012 & HD\,209458   &  85 & 1 & 84.3   & 1 & Rossiter--McLaughlin contamination       & \citet{Laughlin2005} \\
real\_001 & 51\,Peg      & 639 & 1 & 56.0   & 6 & Multi-instrument offsets (6 spectrographs) & \citet{Birkby2017} \\
real\_010 & HD\,189733   &  33 & 1 & 205    & 1 & Active star, sparse data                  & \citet{Boisse2009} \\
real\_009 & HD\,179949   &  88 & 1 & 112.6  & 2 & Hot Jupiter, 2 instruments                & \citet{Butler2006} \\
real\_014 & HD\,217107   & 207 & 1 & 139.7  & 2 & Moderate eccentricity, 2 instruments       & \citet{Wright2009} \\
real\_020 & HD\,88133    &  21 & 1 & 35.7   & 1 & Sparse data (21 points)                   & \citet{Butler2006} \\
\addlinespace
\multicolumn{8}{l}{\textit{Two-planet systems}} \\
\addlinespace
real\_007 & HD\,12661    & 106 & 2 & 74.4   & 2 & Clear period separation                   & \citet{Wright2009} \\
real\_015 & HD\,37124    &  52 & 2 & 28.5   & 1 & Long-period outer planet                  & \citet{Vogt2005} \\
real\_019 & HD\,74156    &  95 & 2 & 125.0  & 2 & High eccentricities ($e > 0.5$)           & \citet{Naef2004} \\
real\_017 & HD\,45364    &  58 & 2 & 21.2   & 1 & 3:2 mean-motion resonance                 & \citet{Correia2009} \\
real\_013 & HD\,215152   & 373 & 2 & 0.87   & 2 & Sub-m\,s$^{-1}$ signals, instrument offset & \citet{Delisle2018} \\
\addlinespace
\multicolumn{8}{l}{\textit{Three-planet systems}} \\
\addlinespace
real\_018 & HD\,69830    &  74 & 3 & 3.51   & 1 & All $K < 4$\,m\,s$^{-1}$                  & \citet{Lovis2006} \\
real\_016 & HD\,40307    & 129 & 3 & 2.54   & 1 & All $K < 3$\,m\,s$^{-1}$, close periods   & \citet{Mayor2009} \\
real\_011 & HD\,20794    & 187 & 3 & 0.85   & 1 & Sub-m\,s$^{-1}$ signals                   & \citet{Pepe2011} \\
\addlinespace
\multicolumn{8}{l}{\textit{Four-or-more-planet systems}} \\
\addlinespace
real\_004 & GJ\,876      & 162 & 4 & 214.0  & 1 & Laplace resonance chain                   & \citet{Rivera2010} \\
real\_008 & HD\,160691   & 380 & 4 & 37.8   & 3 & 3-instrument compilation, wide $K$ range   & \citet{Benedict2022} \\
real\_002 & 55\,Cnc      &  48 & 2$^\dagger$ & 71.3 & 1 & Complex architecture (5 planets known) & \citet{Naef2004} \\
real\_005 & HD\,10180    & 190 & 5 & 4.54   & 1 & 5 planets, low $K$, close periods          & \citet{Lovis2011} \\
real\_006 & HD\,10180 (full) & 190 & 7 & 4.54 & 1 & 7 planets incl.\ sub-m\,s$^{-1}$ signals & \citet{Lovis2011} \\
real\_003 & GJ\,581      & 119 & 4 & 12.5   & 1 & Contested planets, stellar activity        & \citet{Mayor2009gj581} \\
\bottomrule
\end{tabular}%
}

\vspace{0.3em}

{\footnotesize $^\dagger$The \citet{Naef2004} ELODIE dataset for 55\,Cnc contains 48 observations sufficient to constrain 2 of the 5 known planets.}
\caption{Real-world RV tasks included in \textsc{Stargazer}. $N_\mathrm{obs}$ is the number of observations; $N_\mathrm{pl}$ is the number of known planets in the ground-truth model; $K_\mathrm{max}$ is the semi-amplitude of the dominant planet; $N_\mathrm{inst}$ is the number of instruments. Tasks are presented to the agent with anonymised identifiers and no prior information about the planetary system.}
\label{tab:real_data_tasks}
\end{table*}

\paragraph{Data provenance.}
Radial velocity time series were obtained from two primary sources:
the NASA Exoplanet Archive\footnote{\url{https://exoplanetarchive.ipac.caltech.edu/}}~\citep{Butler2006, Wright2009, Lovis_2006}
and VizieR\footnote{\url{https://vizier.cds.unistra.fr/}}~\citep{Naef2004, Mayor2009, Lovis2011}.
Table~\ref{tab:data_provenance} lists the exact archive identifier for each system.
Ground-truth orbital parameters are taken from the discovery or characterisation papers listed in the Reference column.
For multi-instrument datasets, we preserve the original per-instrument RV zero-point offsets; the agent must independently determine and fit these offsets.

\begin{table*}[!htbp]
\centering
\small
\begin{tabular}{cl ll}
\toprule
\textbf{ID} & \textbf{System} & \textbf{Archive} & \textbf{Identifier / Catalogue} \\
\midrule
real\_001 & 51\,Peg        & VizieR & \texttt{J/AJ/153/138} \\
real\_002 & 55\,Cnc        & VizieR & \texttt{J/A+A/414/351} (table3) \\
real\_003 & GJ\,581        & VizieR & \texttt{J/A+A/507/487} (table1) \\
real\_004 & GJ\,876        & VizieR & \texttt{J/ApJ/719/890} (table1) \\
real\_005 & HD\,10180 (5p) & VizieR & \texttt{J/A+A/528/A112} (table1) \\
real\_006 & HD\,10180 (7p) & VizieR & \texttt{J/A+A/528/A112} (table1) \\
real\_007 & HD\,12661      & NASA Exoplanet Archive & UID 0009683 \\
real\_008 & HD\,160691     & VizieR & \texttt{J/AJ/163/295} (table2) \\
real\_009 & HD\,179949     & NASA Exoplanet Archive & UID 0094645 \\
real\_010 & HD\,189733     & VizieR & \texttt{J/A+A/495/959} (table1) \\
real\_011 & HD\,20794      & VizieR & \texttt{J/A+A/534/A58} (table1) \\
real\_012 & HD\,209458     & NASA Exoplanet Archive & UID 0108859 \\
real\_013 & HD\,215152     & VizieR & \texttt{J/A+A/614/A133} (harps\_a, harps\_b) \\
real\_014 & HD\,217107     & NASA Exoplanet Archive & UID 0113421 \\
real\_015 & HD\,37124      & NASA Exoplanet Archive & UID 0026381 \\
real\_016 & HD\,40307      & VizieR & \texttt{J/A+A/493/639} (table1) \\
real\_017 & HD\,45364      & VizieR & \texttt{J/A+A/496/521} (hd45364) \\
real\_018 & HD\,69830      & NASA Exoplanet Archive & UID 0040693 \\
real\_019 & HD\,74156      & VizieR & \texttt{J/A+A/414/351} (table2) \\
real\_020 & HD\,88133      & NASA Exoplanet Archive & UID 0049813 \\
\bottomrule
\end{tabular}
\caption{Data provenance for all 20 real-world RV tasks. NASA Exoplanet Archive entries are identified by their unique dataset ID (UID); VizieR entries are identified by their catalogue designation.}
\label{tab:data_provenance}
\end{table*}

\paragraph{Anonymisation protocol.}
To prevent data contamination from LLM training corpora, each task is assigned an opaque identifier (\texttt{real\_001} through \texttt{real\_020}).
All metadata that could reveal the target identity --- star name, instrument names, literature references, known orbital parameters, and descriptive text --- is stripped from the task file.
Instrument labels are replaced with generic identifiers (\texttt{inst\_A}, \texttt{inst\_B}, \ldots).
The agent receives only: (i)~the time series $(t_i, v_i, \sigma_i)$, (ii)~anonymised instrument labels, and (iii)~the host star mass $M_\star$.

\paragraph{Ground truth and evaluation.}
Real-data tasks are evaluated with the same four criteria as synthetic tasks (\S\ref{sec:evaluation}).
Ground-truth parameters are taken from the peer-reviewed papers listed in Table~\ref{tab:real_data_tasks}; we independently verified literature consistency by re-fitting each system with \texttt{RadVel}.
For contested detections (e.g., GJ\,581\,d/g), we adopt the most widely accepted published solution.

\paragraph{Special case: GJ\,876 (real\_004).}
GJ\,876 hosts a four-planet Laplace resonance chain whose strong planet--planet interactions invalidate the Keplerian superposition assumption; its ground truth is taken from the $N$-body solution of \citet{Rivera2010}.
We retain this system intentionally: it tests whether agents can recognise when the standard fitting workflow breaks down.
In practice, several agents note that a dominant ${\sim}61$\,d signal leaves a stubborn ${\sim}30$\,d residual, and some speculate about resonance or interactions, but none diagnoses the model family itself as misspecified---they continue escalating within the Keplerian search loop rather than switching to a dynamical model.

Table~\ref{tab:real_data_near_matches} summarises the five agent submissions that came closest to the published real-data solutions under our literature-consistency check.

\begin{table*}[!htbp]
\centering
\small
\begin{tabularx}{\textwidth}{l l l c c X}
\toprule
\textbf{Task} & \textbf{Model} & \textbf{Run} & \textbf{$P$ Check} & \textbf{$K$ Check} & \textbf{Main Mismatch} \\
\midrule
real\_018 & GPT-5.2 & Run 1 & all matched planets pass & fail & HD\,69830 b/c/d recovered at $P \approx 8.666, 31.559, 197.330$\,d, but with inflated semi-amplitudes: $K = 8.30/7.23/2.87$\,m\,s$^{-1}$ versus literature $3.51/2.66/2.20$\,m\,s$^{-1}$. \\
real\_018 & Claude-Sonnet-4.6 & Run 2 & all matched planets pass & fail & Same qualitative solution as GPT-5.2 on HD\,69830, with all three periods matching the literature system but all three $K$ values overestimated by several $\sigma$. \\
real\_018 & Gemini-3.1-Pro & Run 2 & all matched planets pass & fail & Again recovers the HD\,69830 three-planet period structure accurately, but returns $K = 8.30/7.23/2.87$\,m\,s$^{-1}$ scale amplitudes rather than the published lower-amplitude solution. \\
real\_020 & Gemini-3.1-Pro & Run 2 & pass & fail & HD\,88133\,b is recovered at the correct short period ($P \approx 3.416$\,d), but the submitted semi-amplitude is $K = 77.6$\,m\,s$^{-1}$ versus the literature value of $36.1$\,m\,s$^{-1}$. \\
real\_018 & GPT-5.3-codex & Run 3 & all matched planets pass & fail & Recovers the HD\,69830 periods, but still overestimates all three amplitudes ($K = 6.65/7.82/3.42$\,m\,s$^{-1}$) relative to the literature solution. \\
\bottomrule
\end{tabularx}
\caption{Closest literature matches among agent submissions on real-data tasks. No submission across the three evaluation runs simultaneously matched all literature periods and semi-amplitudes within the reported uncertainties. The five entries shown here are the closest cases: all matched planets have literature-consistent periods, but at least one semi-amplitude $K$ remains outside the published uncertainty interval.}
\label{tab:real_data_near_matches}
\end{table*}

\section{Per-Criterion Analysis}
\label{app:criterion_analysis}

\subsection{Pass Rate Breakdown by Criterion}
\label{app:per_criterion}

\begin{table*}[h!]
\centering
\resizebox{\textwidth}{!}{%
\begin{tabular}{l c cccc c cccc c cccc}
\toprule
& & \multicolumn{4}{c}{\textbf{Easy}} & & \multicolumn{4}{c}{\textbf{Medium}} & & \multicolumn{4}{c}{\textbf{Hard}} \\
\cmidrule(lr){3-6} \cmidrule(lr){8-11} \cmidrule(lr){13-16}
\textbf{Model} & & \cellcolor{stat-bg}$\Delta$BIC & \cellcolor{stat-bg}RMS & \cellcolor{phys-bg}Match & \cellcolor{phys-bg}Count & & \cellcolor{stat-bg}$\Delta$BIC & \cellcolor{stat-bg}RMS & \cellcolor{phys-bg}Match & \cellcolor{phys-bg}Count & & \cellcolor{stat-bg}$\Delta$BIC & \cellcolor{stat-bg}RMS & \cellcolor{phys-bg}Match & \cellcolor{phys-bg}Count \\
\midrule
Kimi-K2.5       && \cellcolor{stat-bg}100 & \cellcolor{stat-bg}50.0 & \cellcolor{phys-bg}50.0 & \cellcolor{phys-bg}100  && \cellcolor{stat-bg}89.8 & \cellcolor{stat-bg}72.7 & \cellcolor{phys-bg}27.3 & \cellcolor{phys-bg}77.3 && \cellcolor{stat-bg}84.9 & \cellcolor{stat-bg}64.2 & \cellcolor{phys-bg} 1.9 & \cellcolor{phys-bg}35.8 \\
Qwen-3.5-Plus   && \cellcolor{stat-bg}90.0& \cellcolor{stat-bg}85.0 & \cellcolor{phys-bg}80.0 & \cellcolor{phys-bg}100  && \cellcolor{stat-bg}87.5 & \cellcolor{stat-bg}87.5 & \cellcolor{phys-bg}36.4 & \cellcolor{phys-bg}80.7 && \cellcolor{stat-bg}89.8 & \cellcolor{stat-bg}71.2 & \cellcolor{phys-bg} 0.0 & \cellcolor{phys-bg}27.1 \\
o3-mini         && \cellcolor{stat-bg}98.3& \cellcolor{stat-bg}80.8 & \cellcolor{phys-bg}70.0 & \cellcolor{phys-bg}82.5 && \cellcolor{stat-bg}92.5 & \cellcolor{stat-bg}78.6 & \cellcolor{phys-bg}27.7 & \cellcolor{phys-bg}69.2 && \cellcolor{stat-bg}89.7 & \cellcolor{stat-bg}62.3 & \cellcolor{phys-bg} 0.7 & \cellcolor{phys-bg}27.4 \\
GPT-5.2         && \cellcolor{stat-bg}100 & \cellcolor{stat-bg}93.1 & \cellcolor{phys-bg}86.2 & \cellcolor{phys-bg}100  && \cellcolor{stat-bg}94.0 & \cellcolor{stat-bg}96.4 & \cellcolor{phys-bg}43.4 & \cellcolor{phys-bg}77.1 && \cellcolor{stat-bg}95.2 & \cellcolor{stat-bg}90.5 & \cellcolor{phys-bg}11.9 & \cellcolor{phys-bg}59.5 \\
Claude-Sonnet-4.6&& \cellcolor{stat-bg}97.9& \cellcolor{stat-bg}97.9 & \cellcolor{phys-bg}87.2 & \cellcolor{phys-bg}95.7 && \cellcolor{stat-bg}100  & \cellcolor{stat-bg}98.7 & \cellcolor{phys-bg}46.1 & \cellcolor{phys-bg}82.9 && \cellcolor{stat-bg}96.6 & \cellcolor{stat-bg}96.6 & \cellcolor{phys-bg} 3.4 & \cellcolor{phys-bg}10.3 \\
Gemini-3.1-Pro  && \cellcolor{stat-bg}100 & \cellcolor{stat-bg}95.8 & \cellcolor{phys-bg}91.7 & \cellcolor{phys-bg}100  && \cellcolor{stat-bg}95.5 & \cellcolor{stat-bg}97.3 & \cellcolor{phys-bg}39.6 & \cellcolor{phys-bg}76.6 && \cellcolor{stat-bg}88.8 & \cellcolor{stat-bg}86.7 & \cellcolor{phys-bg} 3.1 & \cellcolor{phys-bg}58.2 \\
GPT-5-mini      && \cellcolor{stat-bg}100 & \cellcolor{stat-bg}92.7 & \cellcolor{phys-bg}83.6 & \cellcolor{phys-bg}92.7 && \cellcolor{stat-bg}93.3 & \cellcolor{stat-bg}94.3 & \cellcolor{phys-bg}39.0 & \cellcolor{phys-bg}73.3 && \cellcolor{stat-bg}93.3 & \cellcolor{stat-bg}84.4 & \cellcolor{phys-bg} 0.0 & \cellcolor{phys-bg}25.6 \\
GPT-5.3-codex   && \cellcolor{stat-bg}94.7& \cellcolor{stat-bg}84.2 & \cellcolor{phys-bg}78.9 & \cellcolor{phys-bg}94.7 && \cellcolor{stat-bg}92.6 & \cellcolor{stat-bg}90.7 & \cellcolor{phys-bg}38.9 & \cellcolor{phys-bg}79.6 && \cellcolor{stat-bg}100  & \cellcolor{stat-bg}100  & \cellcolor{phys-bg}33.3 & \cellcolor{phys-bg}33.3 \\
\bottomrule
\end{tabular}%
}
\caption{Per-criterion pass rates (\%) among tasks with $\geq$1 submission, broken down by difficulty tier.
\colorbox{stat-bg}{$\Delta$BIC} and \colorbox{stat-bg}{RMS} measure \emph{statistical detection};
\colorbox{phys-bg}{Match} and \colorbox{phys-bg}{Count} measure \emph{physical recovery}.}
\label{tab:success_criteria}
\end{table*}

Table~\ref{tab:success_criteria} breaks down the pass rate for each of the four evaluation criteria ($\Delta$BIC, RMS, Match Score, Planet Count) by difficulty tier, revealing several patterns not visible in the aggregate pass rates.

\paragraph{Statistical criteria do not distinguish models.}
On Easy tasks, $\Delta$BIC exceeds nearly 90\% for almost all models and RMS remains uniformly high.
Detecting a periodic signal and achieving a reasonable fit is not a bottleneck; the standard periodogram-to-Keplerian pipeline is well within frontier model capabilities.

\paragraph{Match Score and Planet Count capture distinct failure modes.}
On Hard tasks, Planet Count remains above 25\% for several models (GPT-5.2: 59.5\%, Gemini: 58.2\%), indicating that some agents correctly infer the number of planets but recover inaccurate orbital parameters.
Match Score collapses to single digits for most models ($<$5\%), revealing that the bottleneck is not merely deciding \emph{how many} planets exist, but accurately \emph{characterising} their orbits.

\paragraph{Claude-Sonnet-4.6: best fits, worst planet count.}
Claude achieves the highest statistical rates across all tiers (Hard: 96.6\% $\Delta$BIC, 96.6\% RMS), yet its Hard-tier Planet Count is the lowest of all models (10.3\%).
It excels at fitting within a fixed model but systematically fails at model selection.

\paragraph{Conjunction gate is strict.}
GPT-5.3-codex achieves 100\% on both statistical criteria on Hard tasks, yet its overall Hard pass rate is only 4.2\%, because Match and Count must \emph{both} pass simultaneously.

\paragraph{Implication.}
The core bottleneck is \emph{model selection} (choosing the correct number of planets) and \emph{parameter recovery} (accurately determining orbital elements).
Both require physical reasoning beyond optimisation: knowing when to escalate model complexity, recognising alias periods, and diagnosing whether structured residuals reflect missing planets or correlated noise.

\paragraph{Stricter match score aggregation.}
The current implementation averages $S_{\text{match}}$ over successfully paired planets only; unmatched truth planets (with pairwise distance $d_{ij} > 5$) do not contribute to the score. A stricter formulation would normalize by $|T|$ (the number of true planets) rather than $|\mathcal{M}|$ (the number of matched pairs):
\begin{equation}
    S_{\text{match}} = \frac{1}{|T|} \sum_{(i,j)\in\mathcal{M}} e^{-d_{ij}} - 0.25\left|n_{\text{truth}} - n_{\text{guess}}\right|.
\end{equation}
Under this formulation, a submission that correctly identifies one of three planets cannot pass \texttt{ok\_match} regardless of how well that planet is recovered. We retain the current mean-over-matched formulation as the primary metric because frontier models rarely trigger this edge case under current capability levels, but recommend adopting the stricter variant as models improve and partial-recovery solutions become more common.

\subsection{Output Format Compliance}

As noted in Section~\ref{app:json_repair}, Qwen-3.5-Plus and Kimi-K2.5 frequently produce malformed JSON when submitting analysis results.
Our harness includes a regex-based fallback parser that attempts to repair common formatting errors (e.g., trailing commas, unquoted keys, truncated brackets).
Despite this mitigation, a substantial fraction of malformed outputs remain unrecoverable, continuing to consume step budget without advancing the analysis.
The reported pass rates for these two models therefore reflect performance \emph{after} automated repair; without it, their scores would be even lower.

\subsection{Difficulty Factor Correlations}
\label{app:difficulty_factor_correlations}

\begin{figure}[t]
    \centering
    \includegraphics[width=0.70\linewidth]{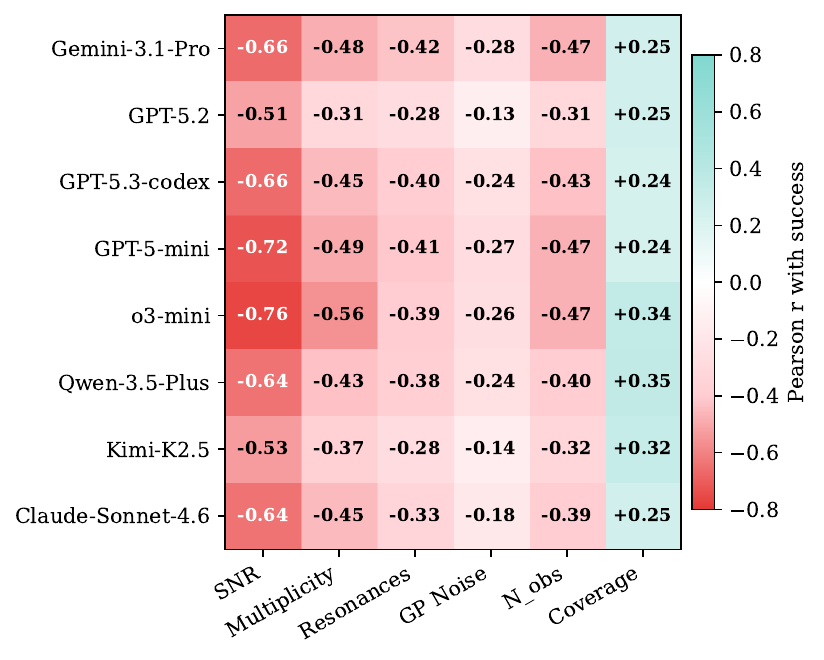}
    \caption{Pearson correlation between difficulty factors and per-task success, aggregated over 100 synthetic tasks and 3 runs per model.}
    \label{fig:failure_correlation}
\end{figure}

Figure~\ref{fig:failure_correlation} reports the Pearson correlation between difficulty factors and per-task binary success (computed over all 100 synthetic tasks and 3 runs per model). Low SNR and higher planet multiplicity are the strongest negative predictors.

\subsection{Per-Criterion Effect of Skills Injection}
\label{app:skills_criteria}

Table~\ref{tab:skills_criteria} decomposes the skills ablation (Section~\ref{sec:skills}) into the four evaluation criteria for all four models.
Three patterns emerge:

\paragraph{Easy tier: uniform procedural improvement.}
All four models show improved or stable $\Delta$BIC, RMS, and Count on Easy tasks after skills injection. Gemini-3.1-Pro reaches 100\% on all four criteria.
This confirms that skills successfully encode the standard periodogram-to-fit workflow.

\paragraph{Medium/Hard: strong models gain on physical criteria.}
The most striking effect is in Match Score on Medium tasks: Gemini-3.1-Pro jumps from 39.3\% to 81.9\% (+42.6\,pp) and GPT-5.3-codex from 40.0\% to 63.8\% (+23.8\,pp).
Planet Count shows similar gains for these two models (+17.9 and +21.4\,pp on Medium).
This indicates that procedural scaffolding helps stronger models land closer to correct orbital solutions, not just achieve better statistical fits.

\paragraph{Weaker models: RMS degrades on Hard tasks.}
GPT-5-mini and Qwen-3.5-Plus show RMS degradation on Hard tasks ($-$12.7 and $-$7.3\,pp respectively), while their Match Scores remain near zero.
This suggests that the procedural template imported from Easy-tier trajectories may actively interfere with the more exploratory fitting strategies required for complex multi-planet systems.

\begin{table*}[t]
\centering
\resizebox{\textwidth}{!}{%
\begin{tabular}{l c cccc c cccc c cccc}
\toprule
& & \multicolumn{4}{c}{\textbf{Easy}} & & \multicolumn{4}{c}{\textbf{Medium}} & & \multicolumn{4}{c}{\textbf{Hard}} \\
\cmidrule(lr){3-6} \cmidrule(lr){8-11} \cmidrule(lr){13-16}
\textbf{Model} & & \cellcolor{stat-bg}$\Delta$BIC & \cellcolor{stat-bg}RMS & \cellcolor{phys-bg}Match & \cellcolor{phys-bg}Count & & \cellcolor{stat-bg}$\Delta$BIC & \cellcolor{stat-bg}RMS & \cellcolor{phys-bg}Match & \cellcolor{phys-bg}Count & & \cellcolor{stat-bg}$\Delta$BIC & \cellcolor{stat-bg}RMS & \cellcolor{phys-bg}Match & \cellcolor{phys-bg}Count \\
\midrule
GPT-5-mini &&
  \cellcolor{stat-bg}100\textsubscript{\textcolor{gray}{\tiny$\cdot$}} &
  \cellcolor{stat-bg}98.3\textsubscript{\textcolor{ForestGreen}{\tiny+2.9}} &
  \cellcolor{phys-bg}90.0\textsubscript{\textcolor{ForestGreen}{\tiny+1.9}} &
  \cellcolor{phys-bg}98.3\textsubscript{\textcolor{ForestGreen}{\tiny+3.8}} &&
  \cellcolor{stat-bg}91.7\textsubscript{\textcolor{gray}{\tiny$\cdot$}} &
  \cellcolor{stat-bg}85.8\textsubscript{\textcolor{red}{\tiny$-$7.9}} &
  \cellcolor{phys-bg}38.3\textsubscript{\textcolor{ForestGreen}{\tiny+2.6}} &
  \cellcolor{phys-bg}79.2\textsubscript{\textcolor{ForestGreen}{\tiny+8.6}} &&
  \cellcolor{stat-bg}94.9\textsubscript{\textcolor{ForestGreen}{\tiny+3.4}} &
  \cellcolor{stat-bg}72.9\textsubscript{\textcolor{red}{\tiny$-$12.7}} &
  \cellcolor{phys-bg}2.5\textsubscript{\textcolor{gray}{\tiny$\cdot$}} &
  \cellcolor{phys-bg}33.9\textsubscript{\textcolor{ForestGreen}{\tiny+2.5}} \\
Qwen-3.5-Plus &&
  \cellcolor{stat-bg}96.1\textsubscript{\textcolor{ForestGreen}{\tiny+3.4}} &
  \cellcolor{stat-bg}92.2\textsubscript{\textcolor{ForestGreen}{\tiny+9.2}} &
  \cellcolor{phys-bg}86.3\textsubscript{\textcolor{ForestGreen}{\tiny+5.8}} &
  \cellcolor{phys-bg}100\textsubscript{\textcolor{gray}{\tiny$\cdot$}} &&
  \cellcolor{stat-bg}89.8\textsubscript{\textcolor{ForestGreen}{\tiny+3.0}} &
  \cellcolor{stat-bg}86.7\textsubscript{\textcolor{gray}{\tiny$\cdot$}} &
  \cellcolor{phys-bg}38.8\textsubscript{\textcolor{ForestGreen}{\tiny+3.6}} &
  \cellcolor{phys-bg}81.6\textsubscript{\textcolor{ForestGreen}{\tiny+2.5}} &&
  \cellcolor{stat-bg}91.5\textsubscript{\textcolor{ForestGreen}{\tiny+5.6}} &
  \cellcolor{stat-bg}66.0\textsubscript{\textcolor{red}{\tiny$-$7.3}} &
  \cellcolor{phys-bg}0.0\textsubscript{\textcolor{red}{\tiny$-$2.8}} &
  \cellcolor{phys-bg}25.5\textsubscript{\textcolor{red}{\tiny$-$4.0}} \\
Gemini-3.1-Pro &&
  \cellcolor{stat-bg}100\textsubscript{\textcolor{gray}{\tiny$\cdot$}} &
  \cellcolor{stat-bg}100\textsubscript{\textcolor{ForestGreen}{\tiny+3.1}} &
  \cellcolor{phys-bg}100\textsubscript{\textcolor{ForestGreen}{\tiny+7.2}} &
  \cellcolor{phys-bg}100\textsubscript{\textcolor{gray}{\tiny$\cdot$}} &&
  \cellcolor{stat-bg}90.4\textsubscript{\textcolor{red}{\tiny$-$5.1}} &
  \cellcolor{stat-bg}93.6\textsubscript{\textcolor{red}{\tiny$-$3.7}} &
  \cellcolor{phys-bg}81.9\textsubscript{\textcolor{ForestGreen}{\tiny+42.6}} &
  \cellcolor{phys-bg}94.7\textsubscript{\textcolor{ForestGreen}{\tiny+17.9}} &&
  \cellcolor{stat-bg}88.7\textsubscript{\textcolor{ForestGreen}{\tiny+2.6}} &
  \cellcolor{stat-bg}87.3\textsubscript{\textcolor{ForestGreen}{\tiny+1.2}} &
  \cellcolor{phys-bg}32.4\textsubscript{\textcolor{ForestGreen}{\tiny+26.5}} &
  \cellcolor{phys-bg}54.9\textsubscript{\textcolor{red}{\tiny$-$7.4}} \\
GPT-5.3-codex &&
  \cellcolor{stat-bg}96.2\textsubscript{\textcolor{gray}{\tiny$\cdot$}} &
  \cellcolor{stat-bg}86.8\textsubscript{\textcolor{gray}{\tiny$\cdot$}} &
  \cellcolor{phys-bg}83.0\textsubscript{\textcolor{red}{\tiny$-$1.2}} &
  \cellcolor{phys-bg}98.1\textsubscript{\textcolor{ForestGreen}{\tiny+3.4}} &&
  \cellcolor{stat-bg}92.4\textsubscript{\textcolor{red}{\tiny$-$2.1}} &
  \cellcolor{stat-bg}84.8\textsubscript{\textcolor{red}{\tiny$-$6.3}} &
  \cellcolor{phys-bg}63.8\textsubscript{\textcolor{ForestGreen}{\tiny+23.8}} &
  \cellcolor{phys-bg}98.1\textsubscript{\textcolor{ForestGreen}{\tiny+21.4}} &&
  \cellcolor{stat-bg}85.1\textsubscript{\textcolor{red}{\tiny$-$14.9}} &
  \cellcolor{stat-bg}75.9\textsubscript{\textcolor{red}{\tiny$-$10.8}} &
  \cellcolor{phys-bg}32.2\textsubscript{\textcolor{red}{\tiny$-$7.8}} &
  \cellcolor{phys-bg}85.1\textsubscript{\textcolor{ForestGreen}{\tiny+31.7}} \\
\bottomrule
\end{tabular}%
}
\caption{Per-criterion pass rates (\%) \emph{with} skills injection, averaged over three runs, among tasks with $\geq$1 submission.
\colorbox{stat-bg}{$\Delta$BIC} and \colorbox{stat-bg}{RMS} are statistical criteria;
\colorbox{phys-bg}{Match} and \colorbox{phys-bg}{Count} are physical criteria.
Superscripts show the change vs.\ the default condition in Table~\ref{tab:success_criteria}
(\textcolor{ForestGreen}{$+$} improvement, \textcolor{red}{$-$} degradation, \textcolor{gray}{$\cdot$} $<$1\,pp).}
\label{tab:skills_criteria}
\end{table*}

\subsection{Episode Termination Under Skills Injection}
\label{app:skills_stop_reasons}

Table~\ref{tab:skills_stop_reasons} decomposes each episode into two mutually exclusive outcomes: the agent finishes naturally (\textbf{Env Done}) or is cut off by a resource limit (\textbf{Budget Exceeded}).

The key observation is that skills shift episodes \emph{from} Budget Exceeded \emph{to} Env Done without changing what the agent does once it reaches the submission stage.
For Gemini-3.1-Pro on Hard tasks, Budget Exceeded drops from 93.5\% to 83.3\% ($-$10.1\,pp), exactly matching the Env Done increase.
This means the pass-rate gains reported in Table~\ref{tab:skills_results} are almost entirely attributable to \emph{efficiency}: skills compress the workflow so that episodes that previously timed out now reach the submission stage.
The underlying physical reasoning is not improved, as Table~\ref{tab:skills_criteria} confirms (Match Score on Hard tasks remains below 33\% for all models).

\providecommand{\tsup}[1]{\textsubscript{\textcolor{ForestGreen}{\tiny+#1}}}
\providecommand{\tsdn}[1]{\textsubscript{\textcolor{red}{\tiny$-$#1}}}
 \begin{table*}[!htbp]
  \centering
  \resizebox{\textwidth}{!}{%
  \begin{tabular}{ll cc cc cc}
  \toprule
  & & \multicolumn{2}{c}{\textbf{Easy}} & \multicolumn{2}{c}{\textbf{Medium}} & \multicolumn{2}{c}{\textbf{Hard}} \\
  \cmidrule(lr){3-4}\cmidrule(lr){5-6}\cmidrule(lr){7-8}
  \textbf{Model} & \textbf{Variant}
    & Pass & Submitted
    & Pass & Submitted
    & Pass & Submitted \\
  \midrule
  \multirow{2}{*}{GPT-5-mini}
    & Default & 76.7 & 76.7  & 31.7 & 48.3  & 2.5 & 5.8 \\
    & +Skills & 90.0\tsup{13.3} & 90.0\tsup{13.3}  & 35.8\tsup{4.1} & 49.2\tsup{0.9}  & 2.5\textsubscript{\textcolor{gray}{\tiny+0.0}} & 5.0\tsdn{0.8} \\
  \addlinespace
  \multirow{2}{*}{Qwen-3.5-Plus}
    & Default & 26.7 & 26.7  & 25.0 & 25.8  & 1.7 & 2.5 \\
    & +Skills & 48.3\tsup{21.6} & 48.3\tsup{21.6}  & 23.3\tsdn{1.7} & 23.3\tsdn{2.5}  & 0.0\tsdn{1.7} & 0.0\tsdn{2.5} \\
  \addlinespace
  \multirow{2}{*}{Gemini-3.1-Pro}
    & Default & 71.7 & 71.7  & 34.2 & 35.0  & 5.0 & 5.8 \\
    & +Skills & 100\tsup{28.3} & 100\tsup{28.3}  & 56.7\tsup{22.5} & 58.3\tsup{23.3}  & 16.7\tsup{11.7} & 16.7\tsup{10.9} \\
  \addlinespace
  \multirow{2}{*}{GPT-5.3-codex}
    & Default & 80.0 & 88.3  & 29.2 & 49.2  & 4.2 & 7.5 \\
    & +Skills & 74.6\tsdn{5.4} & 75.0\tsdn{13.3}  & 53.2\tsup{22.4} & 71.7\tsup{22.5}  & 25.6\tsup{21.4} & 48.3\tsup{40.8} \\
  \bottomrule
  \end{tabular}%
  }
    \caption{Skills injection: Pass Rate vs.\ Submission Rate (\%), averaged over three runs.
  \textbf{Pass} = all four criteria satisfied;
  \textbf{Submitted} = agent produced at least one submission before budget exhaustion.
  Superscripts: \textcolor{ForestGreen}{green} = improvement, \textcolor{red}{red} = degradation vs.\ Default.}
  \label{tab:skills_stop_reasons}
  \end{table*}

\subsection{Match Score Threshold Sensitivity}
\label{app:threshold_sensitivity}

Table~\ref{tab:threshold_sensitivity} reports pass rates under $\pm$10\% variation of the Match Score threshold (default 0.80). The match-score distribution is strongly bimodal: most submissions score either $>$0.9 (correct recovery) or $<$0.5 (wrong parameters), with only 17\% of submissions falling in the 0.70--0.90 boundary region. As a result, threshold variation produces $<$5\,pp change in overall pass rates and preserves relative model rankings across all tiers.

  \begin{table*}[!htbp]
  \centering
  \caption{Sensitivity of pass rate (\%) to the Match Score threshold. The default threshold is 0.80; columns show results under $\pm$10\% variation. Top-tier rankings on
  Easy and Medium tasks are largely preserved; reorderings on Hard tasks reflect near-zero absolute rates where single-task differences dominate.}
  \label{tab:threshold_sensitivity}
  \small
  \begin{tabular}{l ccc ccc ccc}
  \toprule
  & \multicolumn{3}{c}{$S_\mathrm{match} \ge 0.72$ ($-$10\%)}
  & \multicolumn{3}{c}{$S_\mathrm{match} \ge 0.80$ (default)}
  & \multicolumn{3}{c}{$S_\mathrm{match} \ge 0.88$ ($+$10\%)} \\
  \cmidrule(lr){2-4}\cmidrule(lr){5-7}\cmidrule(lr){8-10}
  \textbf{Model}
    & \cellcolor{easy-bg}Easy & \cellcolor{med-bg}Med & \cellcolor{hard-bg}Hard
    & \cellcolor{easy-bg}Easy & \cellcolor{med-bg}Med & \cellcolor{hard-bg}Hard
    & \cellcolor{easy-bg}Easy & \cellcolor{med-bg}Med & \cellcolor{hard-bg}Hard \\
  \midrule
  o3-mini          & \cellcolor{easy-bg}40.5 & \cellcolor{med-bg}27.2 & \cellcolor{hard-bg} 0.5  & \cellcolor{easy-bg}40.0 & \cellcolor{med-bg}24.2 & \cellcolor{hard-bg}
  0.0  & \cellcolor{easy-bg}37.9 & \cellcolor{med-bg}19.9 & \cellcolor{hard-bg} 0.0 \\
  GPT-5-mini       & \cellcolor{easy-bg}81.7 & \cellcolor{med-bg}39.9 & \cellcolor{hard-bg} 3.7  & \cellcolor{easy-bg}76.7 & \cellcolor{med-bg}33.6 & \cellcolor{hard-bg}
  2.5  & \cellcolor{easy-bg}76.7 & \cellcolor{med-bg}28.2 & \cellcolor{hard-bg} 0.0 \\
  GPT-5.2          & \cellcolor{easy-bg}45.0 & \cellcolor{med-bg}39.0 & \cellcolor{hard-bg} 5.8  & \cellcolor{easy-bg}40.0 & \cellcolor{med-bg}30.0 & \cellcolor{hard-bg}
  5.8  & \cellcolor{easy-bg}40.0 & \cellcolor{med-bg}29.9 & \cellcolor{hard-bg} 0.0 \\
  Kimi-K2.5        & \cellcolor{easy-bg}15.0 & \cellcolor{med-bg}20.7 & \cellcolor{hard-bg} 0.8  & \cellcolor{easy-bg}13.3 & \cellcolor{med-bg}17.9 & \cellcolor{hard-bg}
  0.8  & \cellcolor{easy-bg}13.3 & \cellcolor{med-bg}15.1 & \cellcolor{hard-bg} 0.8 \\
  Qwen-3.5-Plus    & \cellcolor{easy-bg}28.4 & \cellcolor{med-bg}31.4 & \cellcolor{hard-bg} 1.6  & \cellcolor{easy-bg}26.7 & \cellcolor{med-bg}25.0 & \cellcolor{hard-bg}
  1.6  & \cellcolor{easy-bg}26.7 & \cellcolor{med-bg}22.2 & \cellcolor{hard-bg} 0.0 \\
  Gemini-3.1-Pro   & \cellcolor{easy-bg}76.7 & \cellcolor{med-bg}40.9 & \cellcolor{hard-bg} 8.1  & \cellcolor{easy-bg}71.7 & \cellcolor{med-bg}35.0 & \cellcolor{hard-bg}
  5.0  & \cellcolor{easy-bg}70.0 & \cellcolor{med-bg}29.9 & \cellcolor{hard-bg} 0.0 \\
  Claude-Sonnet-4.6& \cellcolor{easy-bg}73.4 & \cellcolor{med-bg}26.4 & \cellcolor{hard-bg} 0.8  & \cellcolor{easy-bg}68.3 & \cellcolor{med-bg}22.5 & \cellcolor{hard-bg}
  0.8  & \cellcolor{easy-bg}68.3 & \cellcolor{med-bg}16.7 & \cellcolor{hard-bg} 0.0 \\
  GPT-5.3-codex    & \cellcolor{easy-bg}81.7 & \cellcolor{med-bg}38.0 & \cellcolor{hard-bg} 4.2  & \cellcolor{easy-bg}80.0 & \cellcolor{med-bg}30.8 & \cellcolor{hard-bg}
  4.2  & \cellcolor{easy-bg}78.3 & \cellcolor{med-bg}24.6 & \cellcolor{hard-bg} 0.0 \\
  \bottomrule
  \end{tabular}
  \end{table*}

\section{Domain-Expert Skills}
\label{app:skills_content}

In our skills-injection experiments (\S\ref{sec:skills}), each agent receives five domain-expert skills appended to its system prompt.
Below we reproduce the full text of each skill exactly as provided to the agent.

\begin{tcolorbox}[
  enhanced, breakable,
  colback=skillboxbg, colframe=skillboxframe,
  colbacktitle=skillboxtitle, coltitle=white,
  title={\textbf{Skill 1: RV Period Search \& Alias Detection}},
  fonttitle=\small\bfseries,
  left=6pt, right=6pt, top=4pt, bottom=4pt
]
\footnotesize
\textbf{Description.} Use this skill when searching for planetary orbital periods in RV data, especially to avoid picking up 1-day aliases or harmonics instead of the true period.

\textbf{When to Activate.}
\begin{itemize}
  \item Running a periodogram (GLS, Lomb-Scargle) on RV data
  \item Choosing between multiple peaks in a periodogram
  \item Unsure whether a period candidate is real or an alias
  \item After fitting a planet and residual RMS is unexpectedly high
\end{itemize}

\textbf{Instructions.}
\begin{enumerate}
  \item \textbf{Always detrend first.} Remove linear or polynomial RV drift before running any periodogram. Failure to detrend causes spurious long-period peaks.
  \item \textbf{Reject periods $>$ baseline.} If any periodogram peak has period $>$ (max\_time $-$ min\_time), it is almost certainly a trend alias. Reject immediately.
  \item \textbf{Identify the 1-day alias family.} For candidate $P$, compute: $1/(1/P - 1)$, $1/(1/P + 1)$, $P/2$, $2P$. If a secondary peak lies at one of these, pick the one with higher power and shorter period.
  \item \textbf{Validate with phase-folding.} Phase-fold at top-3 candidates. Random scatter = wrong period; smooth curve = correct period.
  \item \textbf{Narrow refinement.} Refine with fine grid search around $\pm5\%$ of best candidate.
\end{enumerate}
\end{tcolorbox}

\begin{tcolorbox}[
  enhanced, breakable,
  colback=skillboxbg, colframe=skillboxframe,
  colbacktitle=skillboxtitle, coltitle=white,
  title={\textbf{Skill 2: Robust Keplerian Orbit Fitting}},
  fonttitle=\small\bfseries,
  left=6pt, right=6pt, top=4pt, bottom=4pt
]
\footnotesize
\textbf{Description.} Fit a full Keplerian orbit model to RV data. Ensures eccentricity is properly optimized, avoids local minima, and produces reliable parameter estimates.

\textbf{When to Activate.}
\begin{itemize}
  \item After identifying a candidate period from a periodogram
  \item When a sine-fit gives poor RMS or implausible eccentricity
  \item When fitting 1 or more planets to RV data
\end{itemize}

\textbf{Instructions.}
\begin{enumerate}
  \item \textbf{Never use sine-fit as final answer.} Sine fitting assumes $e=0$. Always follow up with a full Keplerian fit with 6 parameters: $P, K, e, \omega, M_0, \gamma$.
  \item \textbf{Use global optimization.} Local optimizers get stuck in local minima. Use \texttt{differential\_evolution} with bounds: $P \pm 2\%$ of periodogram peak, $K \in [0.1, 3K_\mathrm{sine}]$, $e \in [0, 0.8]$, $\omega, M_0 \in [0, 2\pi]$.
  \item \textbf{Polish with local optimizer.} After global search, refine with Nelder-Mead (\texttt{maxiter=10000}).
  \item \textbf{Validate eccentricity.} If $e > 0.8$: suspect artifact, re-fit with $e \leq 0.7$. If RMS with $e=0$ is within 5\% of best-fit RMS: submit circular orbit.
  \item \textbf{Per-instrument offsets.} If multiple instruments, fit independent $\gamma_i$ for each.
  \item \textbf{Report RMS.} If RMS $\gg \sigma_\mathrm{median}$, try different period or starting eccentricity.
\end{enumerate}

\textbf{Common Mistakes.}
Do not submit a sine-fit directly. Do not fix $e=0$ unless justified. Do not use only local optimization. Do not use $t=0$ as reference---always use $t_\mathrm{ref} = \texttt{times[0]}$.
\end{tcolorbox}

\begin{tcolorbox}[
  enhanced, breakable,
  colback=skillboxbg, colframe=skillboxframe,
  colbacktitle=skillboxtitle, coltitle=white,
  title={\textbf{Skill 3: Mean Longitude ($l_\mathrm{rad}$) Calculation}},
  fonttitle=\small\bfseries,
  left=6pt, right=6pt, top=4pt, bottom=4pt
]
\footnotesize
\textbf{Description.} Correctly compute the mean longitude $l_\mathrm{rad}$ at the reference epoch $t_\mathrm{ref} = \texttt{times[0]}$, required for \texttt{submit\_action}. This is the most common source of \texttt{ok\_match} failure.

\textbf{The correct formula:}
\begin{verbatim}
l_rad = (Omega_rad + omega_rad + M0_at_tref) mod 2*pi
\end{verbatim}
For RV-only fits (no astrometry), $\Omega = 0$, so:
\begin{verbatim}
l_rad = (omega_rad + M0_at_tref) % (2 * np.pi)
\end{verbatim}

\textbf{Computing $M_0$ at $t_\mathrm{ref}$.} If your optimizer gives $M_0$ at some time $t_\mathrm{fit}$, convert:
\begin{verbatim}
M0_at_tref = (M0_at_tfit + 2*pi/P * (t_ref - t_fit)) % (2*pi)
\end{verbatim}
If $M_0$ was defined directly at \texttt{times[0]} during fitting (recommended), then $M_0 = M_{0,\mathrm{tref}}$ directly.

\textbf{Sanity checks.}
(1) $l_\mathrm{rad} \in [0, 2\pi]$.
(2) Verify RV at $t_\mathrm{ref}$ matches the first observed RV.
(3) Compute $l_\mathrm{rad}$ independently for each planet.

\textbf{Critical:} Always use \texttt{times[0]} as reference epoch. Using $t=0$, JD\,2450000, or the midpoint will cause \texttt{ok\_match} failure even if all other parameters are correct.
\end{tcolorbox}

\begin{tcolorbox}[
  enhanced, breakable,
  colback=skillboxbg, colframe=skillboxframe,
  colbacktitle=skillboxtitle, coltitle=white,
  title={\textbf{Skill 4: Multi-Planet Detection via Iterative Residual Analysis}},
  fonttitle=\small\bfseries,
  left=6pt, right=6pt, top=4pt, bottom=4pt
]
\footnotesize
\textbf{Description.} Determine the correct number of planets by iteratively fitting and subtracting signals. Multi-planet tasks are the most common cause of \texttt{ok\_count} failure.

\textbf{When to Activate.}
\begin{itemize}
  \item After fitting a first planet and computing residuals
  \item When task difficulty tier is medium to hard.
  \item When residual RMS after first planet fit is still $> 2\times$ the noise floor
\end{itemize}

\textbf{Instructions.}
\begin{enumerate}
  \item \textbf{Check residuals.} Compute residual RMS after subtracting planet 1. If RMS $> 2\times \sigma_\mathrm{median}$, a second planet likely exists. If RMS $\leq 1.5\times \sigma_\mathrm{median}$, likely single-planet.
  \item \textbf{Periodogram on residuals.} Run Lomb-Scargle on residuals, apply alias checks from Skill~1.
  \item \textbf{BIC comparison.} Compute BIC for $N$-planet vs $(N+1)$-planet model. $\Delta\mathrm{BIC} > 10$: strong evidence; $> 6$: moderate evidence; $< 6$: no strong evidence, stop.
  \item \textbf{Joint re-optimization.} After finding approximate $P_2$ from residuals, fit both planets simultaneously.
  \item \textbf{Repeat.} Check 2-planet residuals for a 3rd planet. Stop when residual RMS $\approx$ noise floor or $\Delta\mathrm{BIC} < 6$.
\end{enumerate}

\textbf{Decision Rule:}
Residual RMS $> 3\times$ noise $\to$ add planet;
$2$--$3\times$ noise $\to$ check $\Delta$BIC;
$< 2\times$ noise $\to$ stop.
\end{tcolorbox}

\begin{tcolorbox}[
  enhanced, breakable,
  colback=skillboxbg, colframe=skillboxframe,
  colbacktitle=skillboxtitle, coltitle=white,
  title={\textbf{Skill 5: Submission Strategy \& Timing}},
  fonttitle=\small\bfseries,
  left=6pt, right=6pt, top=4pt, bottom=4pt
]
\footnotesize
\textbf{Description.} Decide when to call \texttt{submit\_action} and avoid analysis paralysis (running out of budget without submitting). No-submission is the worst outcome (reward $= 0$).

\textbf{The Golden Rule: Submit Early, Refine Later.}
Submit a baseline solution as soon as you have a fit with RMS below a reasonable threshold. You can always submit again---only the best submission counts.

\textbf{Submit if ANY of these are true:}
\begin{itemize}
  \item RMS $< 30$\,m/s (for typical tasks)
  \item You have spent $> 8$ tool calls without submitting
  \item Your current fit matches the period and $K$ within 10\%
  \item You are about to try something risky (MCMC, re-fit from scratch)
\end{itemize}

\textbf{Time Budget Allocation.}
\begin{itemize}
  \item \textit{Phase 1---Exploration} (calls 1--8): Load data, periodogram, rough sine fit.
  \item \textit{Phase 2---First Fit} (calls 9--16): Full Keplerian, compute $l_\mathrm{rad}$. $\to$ \textbf{Submit baseline here.}
  \item \textit{Phase 3---Refinement} (calls 17--25): Check residuals for additional planets, refine. $\to$ Submit improved solution if better.
  \item \textit{Phase 4---Polish} (calls 26+): MCMC only if time permits. Do not start from scratch.
\end{itemize}

\textbf{Emergency Protocol.} If close to budget with no submission: immediately fit a circular Keplerian ($e=0$) using the best periodogram period and submit. A rough solution is infinitely better than no submission.
\end{tcolorbox}

\section{System Prompt}
\label{app:system_prompt}

Below is the complete system prompt sent to each agent at the start of a task, shown for a representative Easy-tier task (seed\,44, difficulty\,2).
Dynamic fields (marked with \fbox{boxes}) are populated at runtime from each task's observation data; all other text is shared across tasks.
The prompt specifies the agent's role, available tools, submission format, a mandatory six-step analysis strategy, and common pitfalls.
Budget constraints and fit-quality thresholds scale with task difficulty.

\begin{tcolorbox}[
  enhanced, breakable,
  colback=soft-gray, colframe=soft-line,
  colbacktitle=pantone655, coltitle=white,
  boxrule=1.2pt, arc=4pt,
  fonttitle=\bfseries\large,
  toptitle=4pt, bottomtitle=4pt,
  left=6pt, right=6pt, top=2pt, bottom=2pt,
  title={System Prompt \hfill \normalfont\small\itshape Example: seed\,44 $|$ Easy $|$ 1\,planet}
]
{\scriptsize\ttfamily\setlength{\parskip}{0pt}\setlength{\parindent}{0pt}\setlength{\baselineskip}{7.5pt}%
\newcommand{\secthead}[1]{\vspace{3pt}\par\textsf{\textbf{#1}}\par}%
You are an expert RV data analyst tasked with detecting exoplanets from radial velocity measurements.%
\secthead{\#\#\# Dataset Overview}%
- Number of observations: \fbox{80}\enspace - Time span: \fbox{33.5} days\enspace - RV range: \fbox{-88.10} to \fbox{+74.56} m/s\enspace - Median uncertainty: \fbox{2.60} m/s\quad{\tiny(boxed = dynamic)}%
\secthead{\#\#\# Your Task}%
Analyze the RV data to identify planetary signals. Use the PythonREPL tool to:\par
- Compute periodograms (Lomb-Scargle or other methods)\enspace - Test baseline models (via \texttt{baselines} module)\par
- Fit Keplerian orbital models (NOT simple sinusoids)\enspace - Analyze residuals and trigger optimization when needed%
\secthead{\#\#\# CRITICAL: Keplerian Model Parameters}%
When fitting Keplerian orbits, you MUST fit ALL of these parameters:\par
- \textbf{P}: Period (days)\enspace - \textbf{K}: RV semi-amplitude (m/s)\enspace - \textbf{e}: Eccentricity (0--0.8)\par
- \textbf{omega}: Argument of periastron (radians, 0 to 2pi) --- CRITICAL FOR ECCENTRIC ORBITS!\par
- \textbf{M0}: Mean anomaly at reference time (radians)\enspace - \textbf{gamma}: Systemic velocity offset (m/s)%
\secthead{\#\#\# Submission Format}%
Use Stargazer-native planet fields: P\_days, m\_sin\_i\_mjup, e, omega\_rad, l\_rad\par
l\_rad = mean longitude at reference epoch t\_ref = times\_days[0]. Convert: l\_rad = (Omega\_rad + omega\_rad + M0) \% (2*pi)%
\secthead{\#\#\# Response format for every turn}%
1) Findings: concise hypothesis plus key numbers.\enspace 2) Plan/Next: 1--3 short bullets.\par
3) Code: one fenced code block (only if calling PythonREPL).\enspace 4) Results: outputs interpreted; if ready, submit.%
\secthead{\#\#\# Available Tools}%
1. \textbf{PythonREPL}: Execute Python code. Pre-loaded: times\_days, rvs\_ms, sigmas\_ms, np, baselines, history, star\_mass\_sun, t\_ref\_days, Stargazer\_planet\_from\_fit, Stargazer\_SUBMISSION\_GUIDE\par
2. \textbf{submit\_action}: Submit planet hypotheses (max \fbox{3} planets, period > 0.5\,d, eccentricity 0--0.8)%
\secthead{\#\#\# Budget Constraints}%
- Max tool calls: \fbox{12}\enspace - Max execution time: \fbox{300.0}s\enspace - Max submissions: \fbox{3}\quad{\tiny(scaled by difficulty)}%
\secthead{\#\#\# Mandatory Step 0: Read Protocol Guide First}%
Before any fitting/submission, read Stargazer\_SUBMISSION\_GUIDE in PythonREPL and set: \_protocol\_guide\_ack = True.%
\secthead{\#\#\# Strategy (FOLLOW THIS ORDER)}%
\textbf{Step 1: Periodogram Analysis} --- Compute Lomb-Scargle periodogram; identify strongest peak(s).\par
\textbf{Step 2: Linear Sine Baseline} (MANDATORY) --- Run baselines.baseline\_one\_sine(observation); print period, RMS, RMS/sigma.\par
\textbf{Step 3: Model Gating Decision} --- If RMS/median\_sigma <= 1.5: submit directly; else: escalate to Kepler.\par
\textbf{Step 4: Keplerian Fitting} (ONLY IF GATE=YES) --- Full 6-param fit; multi-start optimisation for e > 0.3.\par
\textbf{Step 5: Check Fit Quality} --- Good fit: RMS approx \fbox{2.60} m/s; bad fit: keep optimising.\par
\textbf{Step 6: Submit ONLY After Convergence} --- Include ALL parameters, especially omega\_rad!%
\secthead{\#\#\# Common Mistakes to AVOID}%
1. Jumping to Kepler before LS + linear-sine gating\enspace 2. Submitting early with poor fit (high RMS)\par
3. Forgetting omega\_rad in submission (defaults to 0!)\enspace 4. Wrong phase convention for l\_rad\par
5. Not doing multi-start optimisation for eccentric orbits%
}
\end{tcolorbox}

\section{Case Study Trajectories}
\label{app:case_studies}

\begin{figure*}[!htbp]
      \centering
      \includegraphics[width=\textwidth]{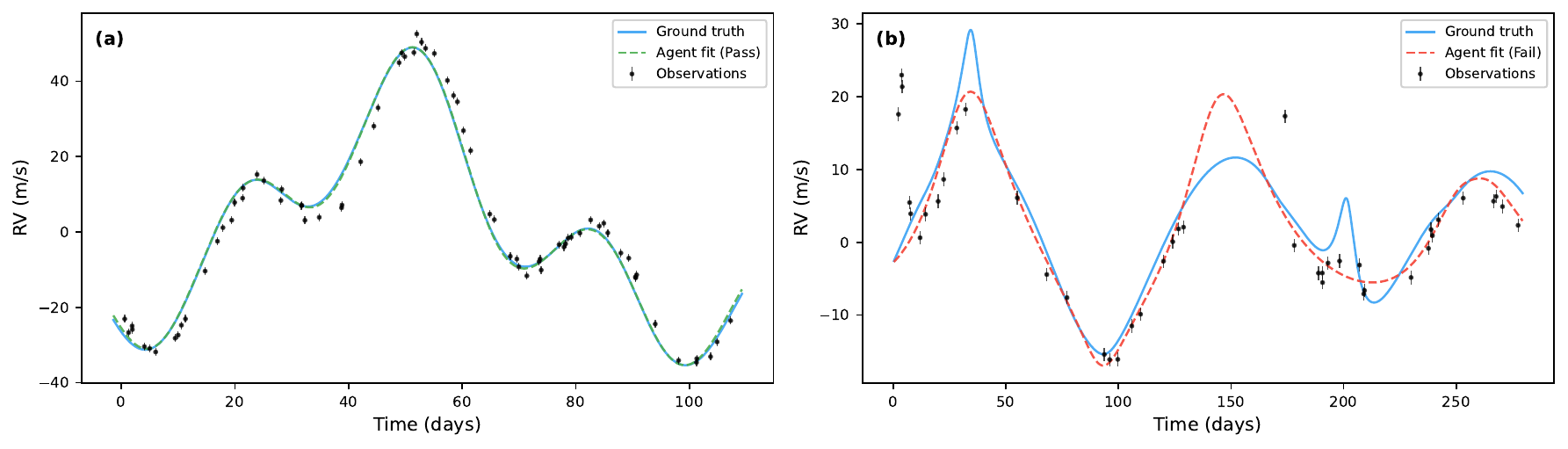}
      \caption{RV model fits for two case studies.
      \textbf{(a)}~GPT-5.2 recovers both planets on seed96 (Medium);
      the agent's fit (green dashed) closely matches the ground truth (blue solid).
      \textbf{(b)}~GPT-5-mini fails on seed196 (Hard):
      its best 2-planet submission converges to alias periods
      ($P = 111.7,\;164.5$\,d instead of the true $76,\;112,\;167$\,d),
      producing an RV curve (red dashed) that visibly diverges from
      the three-planet ground truth.}
      \label{fig:case_study_fits}
  \end{figure*}

Figure~\ref{fig:case_study_fits} shows the RV fits for two representative case studies.

  \begin{tcolorbox}[
    enhanced, breakable,
    colback=white, colframe=soft-line,
    colbacktitle=pantone655, coltitle=white,
    boxrule=1.2pt, arc=4pt,
    fonttitle=\bfseries\large,
    toptitle=4pt, bottomtitle=4pt,
    left=8pt, right=8pt, top=6pt, bottom=6pt,
    title={\cmark\;\;Success Case: 2-Planet Recovery (Medium)
      \hfill \normalfont\small\itshape GPT-5.2\,$|$\,11 steps}
  ]

  {\small
  \begin{tabularx}{\linewidth}{@{}lXlX@{}}
    \toprule
    \multicolumn{4}{@{}l}{\textbf{Task}\enspace Seed \texttt{96}} \\
    \midrule
    \textbf{Planets}       & 2\; ($P = 31.2,\;117.6$\,d) &
    \textbf{Observations}  & 75 \\
    \textbf{Noise model}   & White\; ($\sigma_w = 1.05$\,m\,s$^{-1}$) &
    \textbf{SNR}           & 8.9 \\
    \midrule
    \multicolumn{4}{@{}p{\linewidth}@{}}{%
      \textbf{Difficulty.}\;
      Two-planet system with moderate SNR.
      Planet~2 dominates the RV signal ($K_2 \approx 30$\,m\,s$^{-1}$, $e_2 = 0.25$), but its period ($117.6$\,d) exceeds the data span ($106.7$\,d), causing the initial
  periodogram to converge on a biased estimate ($\sim 105$\,d).
      Planet~1 ($P = 31.2$\,d, $K_1 \approx 14$\,m\,s$^{-1}$) is buried in the residuals of the dominant signal.} \\
    \bottomrule
  \end{tabularx}
  }

  \medskip

  \begin{stepbox}{Steps 1--5 \textmd{\normalfont\ \ Protocol \& Baseline}}
  Reads protocol, computes weighted LS periodogram (40\,000 trial frequencies).
  Dominant peak near $P \approx 105.0$\,d (power $= 0.800$).
  Baseline one-sine fit: $P = 104.8$\,d, $K = 20.3$\,m\,s$^{-1}$, RMS $= 10.68$\,m\,s$^{-1}$ (RMS$/\tilde\sigma = 10.1$).
  Gate decision: \textbf{Kepler = YES.}
  \end{stepbox}

  \smallskip

  \begin{stepbox}{Steps 6--9 \textmd{\normalfont\ \ Iterative Keplerian Fitting}}
  \emph{1-planet fit (multi-start):} converges to $P = 300$\,d at the search boundary---a clear sign the true period exceeds the data span.
  RMS $= 9.16$\,m\,s$^{-1}$.
  Residual periodogram reveals a strong second signal at $P \approx 31.0$\,d (power $= 0.842$).

  \emph{2-planet fit (multi-start):}
  $P_1 = 116.6$\,d, $K_1 = 29.3$\,m\,s$^{-1}$, $e_1 = 0.25$;\;
  $P_2 = 31.1$\,d, $K_2 = 14.0$\,m\,s$^{-1}$, $e_2 = 0.02$.
  RMS: $9.16 \to \mathbf{1.13}$\,m\,s$^{-1}$ (RMS$/\tilde\sigma = 1.07$).
  \end{stepbox}

  \smallskip

  \begin{successbox}{Step 11 \textmd{\normalfont\ \ Submit \#1 (2 planets) $\to$ Pass}}
  Match $= 0.932$,\ RMS $= 1.13$\,m\,s$^{-1}$,\ $\Delta$BIC/pt $= 513$.
  \quad \diagpass{bic}\;
  \diagpass{rms}\;
  \diagpass{match}\;
  \diagpass{count}
  \end{successbox}

  \smallskip
  {\small\textbf{Resources:}\; 61\,K input + 7\,K output tokens, 0 code errors, 1 submission.}

  \begin{insightbox}
  The agent follows a textbook \emph{peel-and-search} strategy: fit the dominant signal, subtract, search residuals for the next.
  A key diagnostic moment occurs at Step~7: the 1-planet fit converges to the period upper bound ($P = 300$\,d), prompting the agent to recognise incomplete convergence and
   check residuals rather than submit.
  The strong residual peak at 31\,d guides the 2-planet fit, which simultaneously refines $P_1$ from the boundary value to the correct $116.6$\,d.
  Both planets are recovered in 11 steps with a single submission, consuming only 68\,K tokens---the most efficient multi-planet recovery in the benchmark.
  \end{insightbox}

  \end{tcolorbox}

\begin{tcolorbox}[
  enhanced, breakable,
  colback=white, colframe=soft-line,
  colbacktitle=pantone655, coltitle=white,
  boxrule=1.2pt, arc=4pt,
  fonttitle=\bfseries\large,
  toptitle=4pt, bottomtitle=4pt,
  left=8pt, right=8pt, top=6pt, bottom=6pt,
  title={\xmark\;\;Failure Case: Stuck in a Local Minimum (Hard)
    \hfill \normalfont\small\itshape GPT-5-mini\,$|$ 40 steps}
]

{\small
\begin{tabularx}{\linewidth}{@{}lXlX@{}}
  \toprule
  \multicolumn{4}{@{}l}{\textbf{Task}\enspace Seed \texttt{196}} \\
  \midrule
  \textbf{Planets}       & 3\; ($P = 76,\;112,\;167$\,d) &
  \textbf{Observations}  & 43 \\
  \textbf{Noise model}   & White\; ($\sigma_w = 0.93$\,m\,s$^{-1}$) &
  \textbf{Resonance}     & Double\; ($\approx 3{:}2{:}1$) \\
  \midrule
  \multicolumn{4}{@{}p{\linewidth}@{}}{%
    \textbf{Difficulty.}\;
    Three planets in a double-resonance chain ($\approx 3{:}2{:}1$), producing alias peaks and combination frequencies.
    The strongest periodogram peak ($\sim 117$\,d) is shifted from the true $P_2 = 112$\,d by unmodelled planets at 76 and 167\,d.} \\
  \bottomrule
\end{tabularx}
}

\medskip

\begin{stepbox}{Steps 1--12 \textmd{\normalfont\ \ Protocol, Baseline \& Fitting}}
LS periodogram finds $P \approx 117.9$\,d.
Baseline: $P = 116.9$\,d, RMS $= 3.35$\,m\,s$^{-1}$ (RMS$/\tilde\sigma = 3.6$). Gate: \textbf{Kepler = YES.}

1-planet fit (108 multi-starts): $P_1 = 116.11$\,d, $e_1 = 0.183$. RMS $= 3.06$\,m\,s$^{-1}$.
Residual periodogram reveals $P_2 \approx 165.9$\,d.
2-planet fit: $P_1 = 111.73$\,d, $P_2 = 164.52$\,d. RMS $= \mathbf{1.08}$\,m\,s$^{-1}$.

\emph{Neither detected period matches a true planet --- both are alias frequencies.}
\end{stepbox}

\smallskip

\begin{failbox}{Steps 13--16 \textmd{\normalfont\ \ Submissions \#1--\#2 $\to$ Format Errors}}
Submit \#1: \texttt{l\_rad} set to 0 $\Rightarrow$ RMS $= 7.71$\,m\,s$^{-1}$. Reward $= -24.6$.
Submit \#2: wrong \texttt{inc\_rad} $\Rightarrow$ RMS $= 9.39$\,m\,s$^{-1}$. Reward $= -45.3$.
\emph{Two submissions wasted on format debugging before the model itself is evaluated.}
\end{failbox}

\smallskip

\begin{failbox}{Steps 17--20 \textmd{\normalfont\ \ Submit \#3 (correct format) $\to$ Alias Fail}}
Properly formatted submission evaluates at RMS $= 1.08$\,m\,s$^{-1}$ (passes \texttt{ok\_rms}),
but \textbf{match $= -0.008$}: the detected periods $(111.7, 164.5)$\,d are alias periods
of the true near-resonant system. Reward $= 18.2$.
\end{failbox}

\smallskip

\begin{failbox}{Steps 21--40 \textmd{\normalfont\ \ Submissions \#4--\#10 $\to$ Perseveration}}
Seven further submissions: the agent resubmits the same 2-planet alias solution 4 times identically (reward $= 18.2$),
tries a 1-planet fallback (reward $= 11.9$), and attempts minor parameter variations.
\textbf{Never attempts a 3-planet model} despite clear residual structure.
No new period candidates are explored in the final 20 steps.
\end{failbox}

\smallskip
{\small\textbf{Resources:}\; 696\,K input + 34\,K output tokens, 14 code errors, 10 submissions --- all failed.}

\noindent{\small\textbf{Diagnostics:}\;
\diagpass{bic}\;
\diagpass{rms}\;
\diagfail{match}\;
\diagfail{count (2 vs.\ 3 true)}}

\begin{insightbox}
Three compounding failure modes:
\emph{(1)~Format fragility}: 30\% of submission budget wasted on API errors (\texttt{l\_rad}, \texttt{inc\_rad}).
\emph{(2)~Alias convergence}: the detected periods $(111.7, 164.5)$\,d are combination frequencies of the true $3{:}2{:}1$ resonant system --- a classic RV pitfall where aliased periods produce low RMS but match $\approx 0$.
\emph{(3)~Perseveration}: the same wrong answer is resubmitted 4 times without model escalation.
The failure consumes $10.7\times$ more tokens (730\,K vs.\ 68\,K) than the success case yet produces a negative match
score, demonstrating that more computation does not compensate for the inability to revise the
model hypothesis.
\end{insightbox}

\end{tcolorbox}

\end{document}